\documentclass[10pt,twocolumn,letterpaper]{article}

\usepackage{iccv}
\usepackage{times}
\usepackage{epsfig}
\usepackage{graphicx}
\usepackage{rotating}
\usepackage{amsmath}
\usepackage{amssymb}
\usepackage{bbm}
\usepackage{caption}
\usepackage{subcaption}
\usepackage{siunitx}
\usepackage{multicol}

\usepackage{pdflscape}

\usepackage[nolist]{acronym}
\usepackage[inline]{enumitem}

\usepackage{siunitx}  
\usepackage{xcolor}

\usepackage[noadjust]{cite}

\DeclareCaptionLabelSeparator{periodspace}{.\quad}
\newcommand\tus{tus_po_8p_dn19_up/qeval/test}
\newcommand\argo{argo_po_8p_dn19_up/qeval/val} 
\newcommand\aerial{aerial_po_8p_dn19_up/qeval/val}

\newcommand{\first}[1]{\textbf{#1}} %
\newcommand{\snd}[1]{\underline{#1}}

\newcommand{\norm}[1]{\left\lVert#1\right\rVert}

\usepackage[accsupp]{axessibility}  %

\usepackage{footnote}
\makesavenoteenv{tabular}
\makesavenoteenv{table}
\usepackage[breaklinks=true,letterpaper=true,bookmarks=false]{hyperref}

\usepackage[nameinlink]{cleveref}

\iccvfinalcopy %

\def\iccvPaperID{****} %

\begin{acronym}
	\acro{lreppo}[Po]{unbound Cartesian points}
	\acro{lrep1d}[1D]{1D border points}
	\acro{lrepeu}[Eu]{Euler angles}
	\acro{MSE}[MSE]{mean squared error}
	\acro{BEV}[BEV]{bird's eye view}
	\acro{KAI}[KAI]{Karlsruhe Aerial Images}
	\acro{CNN}[CNN]{convolutional neural network}
	\acro{NMS}[NMS]{non-maximum suppression}
	\acro{LSD}[LSD]{line segment detection}
	\acro{RNN}{recurrent neural network}
\end{acronym}
\ificcvfinal\fi

\begin{document}

\title{YOLinO: Generic Single Shot Polyline Detection in Real Time}

\author{Annika Meyer
\and 
Philipp Skudlik\thanks{equally contributed} 
\and
Jan-Hendrik Pauls\footnotemark[1] 
\and 
Christoph Stiller
\and 
Institute of Measurement \& Control Systems, Karlsruhe Institute of Technology\\
{\tt\small annika.meyer@kit.edu} 
}

\maketitle
\ificcvfinal\thispagestyle{empty}\fi

\begin{abstract}
The detection of polylines is usually either bound to branchless polylines or formulated in a recurrent way, prohibiting their use in real-time systems. 

We propose an approach that builds upon the idea of single shot object detection. 
Reformulating the problem of polyline detection as a bottom-up composition of small line segments allows to detect bounded, dashed and continuous polylines with a single head.
This has several major advantages over previous methods.
Not only is the method at 187 fps more than suited for real-time applications with virtually any restriction on the shapes of the detected polylines. 
By predicting multiple line segments for each cell, even branching or crossing polylines can be detected.

We evaluate our approach on three different applications for road marking, lane border and center line detection.
Hereby, we demonstrate the ability to generalize to different domains as well as both implicit and explicit polyline detection tasks.

 \end{abstract}

\section{Introduction}

\begin{figure}[t]
	\begin{center}	
	    
				\includegraphics[width=1.\linewidth,trim=0 224 0 0, clip]{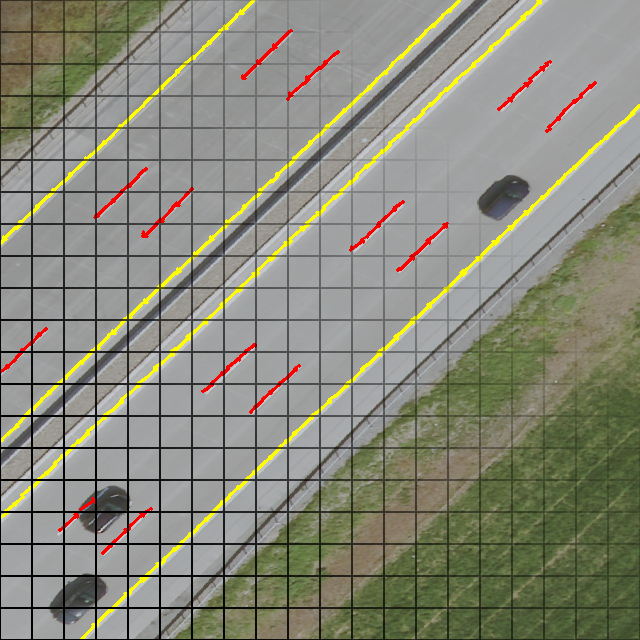}
        \caption{YOLinO detects generic polylines as a bottom-up composition.
		 The image is split into spatial cells each predicting multiple line segments. As one advantage, dashed and solid road markings can be detected using the same head.
		 Aerial image: \textcopyright~City of Karlsruhe $\vert$ Liegenschaftsamt}
    	\label{fig:teaser}
    \end{center}
\end{figure}

Polylines are ubiquitous in many applications. 
They enable a generic representation of real world objects like lane markings, curbs etc. 
However, the recognition of polylines in images is usually bound to branchless polylines or formulated in a recurrent way. Moreover, many approaches can only detect a single polyline, while the majority of applications would need a multitude (\eg lane markings in \autoref{fig:teaser}). 

Learning the detection of polylines so far required recurrent architectures, 
because the number of polylines is often unknown and every polyline contains an indefinite number of vertices (\eg \cite{DAGMapper}).
At the same time, recurrent architectures are
difficult to train and are comparably slow.
Feed-forward neural networks, however, need a well-defined rigid output format that seems incompatible with the variable nature of polylines at first sight.

We propose an approach that is inspired by single-shot object detection \cite{yolov1,yolov2} and is able to detect bounded, dashed and continuous polylines in real time without a recurrent architecture.
Instead of recurrent node proposals, we reformulate the problem of polyline detection as bottom-up composition of small line segments. 
First, the image is fed into a neural network that predicts multiple line segment proposals for independent spatial cells (cf.~\autoref{fig:teaser}).
Later, the line segments can be combined into polylines of virtually any shape or topology.

The explicit encoding of start and endpoints allows predicted lines to carry directional information. Furthermore, they can be enhanced with class labels to differentiate between different line types where such a classification is useful (e.g. lane marking types in \autoref{fig:teaser}).
Lastly, we can represent overlapping polylines in the image.
In fact, our parametrized system is able to even estimate polylines that head in opposite directions while occupying the exact same pixel space.
Facilitating polylines for robotic tasks is intuitive as most ground truth data is provided as polylines in modern datasets \cite{Argoverse,pauls2018can}.

\paragraph{Contributions and outline}
We propose a bottom-up polyline detection approach that is inspired by single shot object detection but serves the same needs as recurrent approaches. For the first time, this allows to obtain a polyline detector that is not only easy and stable to train. With an inference speed of 187~fps, it also allows to use polyline detections in real-time applications such as automated driving.
Thereby, we close a gap that is left in previous works which are reviewed in the next section.%

In \Cref{sec:architecture} and \Cref{sec:non_maximum_suppression}, we describe our approach and the corresponding non-maximum suppression, respectively.
Using three datasets around autonomous driving, in \Cref{sec:datasets} and \Cref{sec:experiments},  we demonstrate the generality of our approach for implicitly and explicitly visible polylines that range from short dashes to continuous and even crossing or branching polyline structures.
Still, for the TuSimple benchmark as an exemplary application, we show that our approach can reach the performance of networks that are tailored just to that very task.
\section{Related Work}
\label{sec:related_work}

Related approaches can be separated into three previously unrelated groups. First, we review single shot object detection approaches that our idea is based on. The second group is \acf{LSD} which detects line segments assuming they are straight, clearly delimited and independent of each other. Finally, we present 
recurrent approaches. 

An approach that fits neither category, but has conceptual similarity to our \acl{lrep1d} representation is~\cite{liao2018deep}. Based on points as input, object surfaces are estimated. A transfer to the image domain is generally conceivable, but a suitable encoder is probably as complex as our whole approach. 

\paragraph{Single shot detection} Object detection can mainly be divided into one- and two-stage approaches. However, alternatives like multi-stage approaches~\cite{CascadeRCNN} exist.

Two-stage approaches have a region proposal step that predicts sparse regions of interest. They are usually more accurate, but also more complex and less efficient. Representatives of this group are the anchor-based R-CNN family~\cite{RCNN,FastRCNN,FasterRCNN} and R-FCN~\cite{RFCN}, but there are also anchor-free two-stage approaches~\cite{RepPoints}.

Motivated by real-time capable inference times and simplicity, single shot object detection approaches have been proposed. They skip the region proposal step and densely predict objects directly on the image. Most famously, SSD~\cite{SSD} and the recently extended YOLO family~\cite{yolov1,yolov2,yolov3,YOLOv4}, but also RetinaNet~\cite{RetinaNet} and EfficientDet~\cite{EfficientDet} are based on static anchors. However, as for two-stage approaches, anchor-free, but slower alternatives have been proposed~\cite{CornerNet,CenterNet,FCOS}.

While both, one- and two-stage approaches are continuously being improved, we wanted to build upon a foundation that is both real-time capable and well-known to be applicable to other tasks than object detection. Thus, we evaluated both YOLO9000~\cite{yolov2} and YOLOv3~\cite{yolov3} with YOLOv3 showing no measureable improvement over YOLO9000 for our task. 
For highest performance, we suggest to transfer our idea to recent architectures like YOLOv4 or EfficientDet, but claim that YOLO9000 is sufficient to show the overall idea and give an estimate of its performance. 

\paragraph{Line segment detection} The problem of detecting (parts of) straight lines has been widely covered under the term \acf{LSD}. For most approaches that do not employ deep learning, we refer to the work by von~Gioi~\cite{LSD,OnLineSegments} that is still often used as baseline. In \cite{LSD}, they also provide a well-written and comprehensive review of multiple classical approaches and key ideas. \cite{MCMLSD} and \cite{YorkUrbanLinesDataset} not only proposed the probably most significant improvements compared to~\cite{LSD}, they also provided two of the main datasets for evaluating \ac{LSD} quantitatively, both based on the YorkUrban images~\cite{YorkUrbanDataset}. However, by concept, all those approaches are limited to lines along edges in the image and cannot detect implicit lines, such as conceptually connected dashes or lane center lines.

Given the aforementioned and the Wireframes~\cite{WireframesDataset} dataset, deep learning approaches could tackle the problem of \ac{LSD}. \cite{WireframesDataset} proposed to detect both junction points and lines connecting them. \cite{AFM,YorkUrban} proposed to use attraction fields for \ac{LSD}, significantly improving the state of the art. While those three approaches used heuristics to finally extract line segments, \cite{LCNN} proposed the first end-to-end-trainable approach for \ac{LSD}. Recently, \cite{HAWP} reformulated the idea of attraction fields in an end-to-end trainable fashion.

Still, these approaches focus on more or less explicit lines. Also, in contrast to polyline detection, \ac{LSD} methods generally target wireframes or similar edges that are densely distributed in the image, without any semantic meaning and usually straight. This makes them considerably different from the polylines that we tackle, but a domain transfer could be tried.

\paragraph{Polyline detection} This brings us to approaches that aim for the very same generic polylines or polygons that we are trying to detect. They can be found in two fields: instance segmentation and road network/road boundary extraction from bird's eye view imagery. 

For highly accurate, but automated instance segmentation, a series of \acp{RNN} has been proposed~\cite{PolygonRNN,PolygonRNNPlusPlus}. They each expect a bounding box/crop around an object and then predict the polyline vertices node by node with optional refinement using gated graph neural networks~\cite{PolygonRNNPlusPlus}. Not only are these \acp{RNN} usually slow and difficult to train, they also need special care to predict the initial vertex.
\cite{CurveGCN,PolyTransform} rely on a more generic initialization that fits well for the instance segmentation case, but not at all to our exemplary applications. While being faster than the recurrent approaches, with around \SI{30}{ms} \cite{CurveGCN} is still far away from the speed of single shot detectors. Also, extra steps would need to be added to care about branches and merges in the graph.

For the extraction of road network information, multiple approaches predict a graph vertex by vertex. \cite{RoadTracer} proposed to predict either a direction or a STOP token using a \ac{CNN}. This enables the automatic tracing of the road network given an initial vertex and allows cycles in the network using a distance-based loop closure.
In \cite{DAGMapper}, two \acp{RNN} are predicting both direction and position of the next vertex that leads along the road boundary. In addition, the \ac{RNN} estimates the branching degree for each node, allowing to extract directed acyclic road boundary graphs.
In contrast, \cite{PolyMapper} directly predicts the position of the next vertex using a \ac{RNN}, proposing to convert a graph into multiple bidirectional branchless polygons.
Finally, \cite{NTG} predicts all neighboring nodes with a recurrent decoder and checks for overlap with the existing graph to close loops.

To the best of our knowledge, all polyline detection approaches have in common that they are either recurrent and, thus, hard to train and slow, or very limited in their topology. This motivated our search for a generic, yet fast method to detect polylines of virtually any shape.

\section{Generic polyline detection}
\label{sec:architecture}
\textit{YOLinO} is inspired by the YOLO object detector family~\cite{yolov1,yolov2,yolov3,complexYolo}, but refined for the purpose of polyline detection. The overall idea is to split the image into several spatial cells and predict multiple line proposals for each cell. Thus, we enable real-time capability with a shallow and simple network, while being flexible to generalize to any kind of polyline without any restriction on the shape. 
\paragraph{Architecture}
While any backbone with sufficient downsampling capabilities may be used, we chose YOLO9000~\cite{yolov2}. We also ran experiments with YOLOv3~\cite{yolov3}, but found it to be outperformed by the smaller version when applied to polyline detection.

DarkNet-19 is a backbone consisting of 19 repeated convolutions and five maxpooling layers to produce a fixed size feature map which represents the cell grid. 
It is then expanded with an additional convolution to predict the parameters for a given number of line segments within each cell. The full architecture can be found in the supplementary material.
\paragraph{Predictors}
Each predictor is constructed as $P = (g,l,c)$ with geometric definition $g$, a class confidence $l$ and the confidence of the predictor as $c$.
The number of predictors per cell consequentially puts an upper limit on the number of line segments that can be predicted per cell.

\paragraph{Grid resolution}
In its vanilla version, DarkNet-19 reduces an input image by a factor of 32.
An input image of $640 \times 640$ for example, translates to a grid size of $20 \times 20$ and a cell size of $32\times32$ pixels.
As such, the grid resolution is tightly bound to the network architecture.

As different applications may benefit from finer grid resolutions, we add a custom network layer for rescaling the output of the feature extractor backbone.
This new upsampling block consists of a transposed convolution and three regular convolutional layers.
When upsampling, we also employ skip connections in order to keep fine-grained features in the process. We evaluate three different upsampling configurations resulting in a grid cell of either $32\times32$ pixels, $16\times16$ pixels or $8\times8$ pixels.
\begin{figure}[t]
	\centering
	\begin{subfigure}[t]{0.32\linewidth}
		\includegraphics[width=1\linewidth,trim=8 8 278 8,clip]{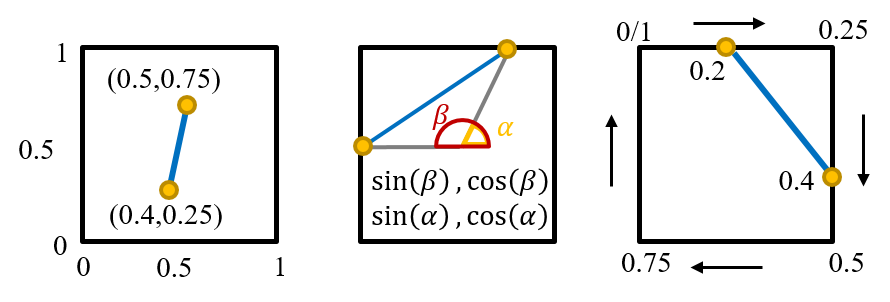} 
		\caption{Cartesian points}
		\label{fig:lreppo}
	\end{subfigure}	
	\begin{subfigure}[t]{0.32\linewidth}
		\includegraphics[width=1\linewidth,trim=288 8 0 8, clip]{figures/ppt_lrep_triple.png}
		\caption{1D points}
		\label{fig:lrep1d}
	\end{subfigure}		
	\begin{subfigure}[t]{0.32\linewidth}	
		\includegraphics[width=1\linewidth,trim=144 8 144 8,clip]{figures/ppt_lrep_triple.png}
		\caption{Euler angles}
		\label{fig:lrepeu}
	\end{subfigure}	
	\caption{Representations for line segments in a spatial cell. Unbound Cartesian Points (\acs{lreppo}) are the most flexible variant, whereas \acf{lrep1d} and \acf{lrepeu} converge faster, although the former suffer from a discontinuity in the top left corner.}	
	\label{fig:lrep}
\end{figure} %

\paragraph{Line representation}
A suitable set of geometric parameters $g$ must be defined to describe the position, dimension and direction of a line segment.
Furthermore, the representation must enable an efficient and meaningful calculation of distances $d(g,\hat{g})$ between line segments for the loss function.
We evaluate several line representations and run extensive evaluations with three variants:
	\acf{lreppo},
	\acf{lrep1d} and
	\acf{lrepeu}.
Note that for all line representations, the start and endpoints are defined by their positions in the network output and, thus, a directional information is encoded implicitly. 

The \acfi{lreppo} define a line segment intuitively with two Cartesian points that are not bound to any restriction in length or position within a grid cell (see \autoref{fig:lreppo}): $g_{\text{Po}} = (g_s, g_e)$ with $g_s,g_e \in \left[0, 1\right]\times\left[0, 1\right]$. We later encountered this concept to be quite similar to the object detection representation presented in \cite{CornerNet}.  

A positional loss can be calculated as distance between two line segments $g$ and $\hat{g}$ with the sum of Euclidean distances between both start and endpoint of the two lines.
\begin{equation} \label{eq:euclidean}
d_{\text{Po}}(g,\hat{g}) = \norm{g_s-\hat{g}_s}_2 + \norm{g_e-\hat{g}_e}_2
\end{equation}
As an alternative, we investigate two other line representations that restrict line segments to have their start and endpoints on the border of a cell.
We expect both to converge faster and limit the search space in order to predict connected polylines. 

The minimal parametric \acfi{lrep1d} regard the cell border as a one-dimensional line, starting in the top left corner and wrapping around the cell clockwise (see \autoref{fig:lrep1d}).
In this setting, a line is described by any tuple of two positions: $g_{\text{1D}} = \left(g_s,g_e\right)$ with $g_s, g_e \in \left[0,1\right]$ along the border.
We calculate the distance between two line segments $g$ and $\hat{g}$ as follows:
\begin{equation} \label{eq:oned}
d_{\text{1D}}(g, \hat{g}) = \sum_{i\in\left\{s,e\right\}} \min(|g_i - \hat{g}_i| , |1 - (|g_i - \hat{g}_i|)|) 
\end{equation}
The \acfi{lrepeu} solve the discontinuity problem of \acfi{lrep1d} at
points close to the top left corner \cite{complexYolo}. 
We represent a line segment as cosine and sine components of the start and endpoint angles, $\alpha$ and $\beta$: $g_{\text{Eu}} = \left(\cos(\alpha),\sin(\alpha),\cos(\beta),\sin(\beta)\right)^\intercal$.

Like for the \acfi{lrep1d}, start and endpoints both lie on the cell border. The angles $\alpha$ and $\beta$ are then measured between the $y$-axis of the image and the line between the cell center and the respective border point~(see \Cref{fig:lrepeu}).
Due to the given periodicity of the representation, the distance is continuously defined between the individual components. 
Similarly to \cite{complexYolo}, we make use of the \ac{MSE} between all components.
\begin{equation}
d_{\text{Eu}}(g, \hat{g}) = \frac{1}{4} \norm{
\begin{pmatrix}
\cos\left(\alpha\right) \\
\sin\left(\alpha\right) \\
\cos\left(\beta\right) \\
\sin\left(\beta\right) \\
\end{pmatrix}
-
\begin{pmatrix}
\cos(\hat{\alpha}) \\
\sin(\hat{\alpha}) \\
\cos(\hat{\beta}) \\
\sin(\hat{\beta}) \\
\end{pmatrix}
}_2
\end{equation}

\paragraph{Training objective}

One of the most crucial ideas to conceptualize when dealing with single shot detection
systems is the notion of responsibility \cite{yolov1}.
When calculating the error term for a given network output, each of the ground truth lines is only matched with a single predictor.
To find the predictor, that is considered responsible for a given ground truth line segment, we opt for an iterative greedy approach: 

Within each cell, we match the closest pair of ground truth line and predicted line until all ground truth lines are assigned a responsible predictor. 
Ultimately, not all predictors will be matched to a ground truth line. 
However, this is intended as each predictor should be responsible only for a specific class/type of line segments that might not be present in the current image.
Our matching process is order-invariant, computationally manageable and enables a responsibility concept for different classes and geometric types of line segments to be learned by different predictors.

We formulate a loss function for this purpose as weighted sum of four sub loss terms. %
These loss terms are defined in \Cref{eq:geometric_loss,eq:conf1,eq:conf2,eq:class_loss}.
\begin{equation}
	\label{eq:geometric_loss}
	\mathcal{L}_{loc} = \sum_{i=0}^{S}\sum_{j=0}^{L}\sum_{k=0}^{P} \mathbbm{1}_{ijk}^{resp} d(g_{ij}, \hat{g}_{ik}) 
\end{equation}
\begin{equation}
	\label{eq:conf1}
	\mathcal{L}_{resp} = \sum_{i=0}^{S}\sum_{k=0}^{P} \mathbbm{1}_{ik}^{resp} (c_{ik} - 1)^2 
\end{equation}
\begin{equation}
	\label{eq:conf2}
	\mathcal{L}_{noresp} = \sum_{i=0}^{S}\sum_{k=0}^{P} (1-\mathbbm{1}_{ik}^{resp}) (c_{ik} - 0)^2
\end{equation}
\begin{equation}
	\label{eq:class_loss}
	\mathcal{L}_{class} = \sum_{i=0}^{S}\sum_{j=0}^{L}\sum_{k=0}^{P} \mathbbm{1}_{ijk}^{resp} \sum_{c=0}^{C}(l_{ik}(c) - \hat{l}_{j}(c))^2  
\end{equation}

Here, $S$ denotes the number of cells within the output grid, $L$ the number of ground truth lines, $P$ the number of predictors and $C$ the number of classes.

\autoref{eq:geometric_loss} defines the loss $\mathcal{L}_{loc}$ on the geometric distance $d(g,\hat{g})$ between two line segments and encapsulates the localization error of all assignments. 
$\mathbbm{1}_{ijk}^{resp}$ indicates if in cell $i$, ground truth line $j$ was assigned to predictor $k$ by the above-mentioned matching process.

\autoref{eq:conf1} and \autoref{eq:conf2} build the confidence error for any predicted line segment.
$c_{ik}$ is the confidence score of the $k$\textsuperscript{th} predictor with the $i$\textsuperscript{th} cell.
$\mathbbm{1}_{ik}^{resp}$ denotes if the $k$\textsuperscript{th} predictor in cell $i$, was assigned to any ground truth line.

\autoref{eq:class_loss} calculates the classification error $\mathcal{L}_{class}$
of the classification $l_{ik}(c)$ for the $k$\textsuperscript{th} ground truth line in cell $i$ for class $c$.

\section{Non-maximum suppression}
\label{sec:non_maximum_suppression}
Since our method outputs a collection of line segments, we propose a simple, yet effective post-processing.
Initially, it is important to set a suitable confidence threshold to keep only relevant predictors.
While the resulting output may already be useful for some downstream applications, the precision can drastically be improved by adding appropriate \ac{NMS}.
The objective of \ac{NMS} is to reduce the number of predictions to the correct density and remove duplicates. While this is rather straight-forward for object detections, the detection of polylines bears some ambiguities. 
Especially for implicit labels like centerlines, many close predictions might be accounted as correct by a human and, thus, manually annotated ground truth labels are noisy by definition.
Furthermore, the neighbor cells which construct a polyline influence the correctness of a predictor and, thus, need to be regarded in the process. 
Hence, for \ac{NMS}, we propose to compare the predictions with a confidence score $c_{ik} > \tau_c$ across multiple spatial cells in the image domain. 

Independent of the chosen line representation, redundant predictions are clustered using DBSCAN~\cite{ester1996density} on $\left(m_x, m_y, l, d_x, d_y\right)^\intercal$ where $m_x, m_y$ are the center point image coordinates, $l$ is the length and $d_x, d_y$ are normalized directions of each predicted line segment. 
We calculate the mean representative of each cluster as result of the \ac{NMS}. 
Here, the confidences are used as weights for both clustering and averaging each cluster. 
This process drastically increases the precision of the estimate while hardly impairing recall and, at 230~fps, not impairing the inference speed. All details on \ac{NMS} can be found in the supplementary material.

\section{Datasets}
\label{sec:datasets}

Our method for image based polyline estimation is designed to be applicable to a wide range of applications.
To prove the method's adaptability, we present results on three different datasets covering two domains and three distinct tasks from photogrammetry and robotics.

\paragraph{TuSimple: Lane boundary estimation}
The TuSimple lane detection benchmark\footnote{\url{https://github.com/TuSimple/tusimple-benchmark/tree/master/doc/lane\_detection}} presents the task of estimating lane boundaries on a highway setting.
The popular dataset features more than $3000$ front view images for training, $358$ for validation and $2782$ for testing.
The polylines are continuous, each describing the boundary between two lanes, indifferent of the type of lane marking.

\paragraph{\acf{KAI}: Lane marking detection}
We use aerial images of a highway section near Karlsruhe, Germany, for which the actual lane markings have been annotated manually as polylines.
This dataset~\cite{pauls2018can} contains more than 1400 \ac{BEV} images for training and the task is to identify the explicit white lane marking instead of the implicit lane boundaries of TuSimple. It is important to note here that the dataset provides both continuous as well as dashed markings, requiring the detection to cope with both continuous and bounded labels as well as a classification of these. 
\paragraph{Argoverse: Lane centerline estimation}
As third dataset, we work with Argoverse~\cite{Argoverse}, using its underlying high definition map information to project lane centerlines into more than 2600 front view images.
This requires the model to estimate \textit{implicit} polylines that do not have direct visual cues within the image.
Given that lane centerlines are often ambiguous and prone to overlaps and occlusions, this problem setting is considered the most difficult of the three.
Unfortunately, there is neither in Argoverse nor anywhere else a benchmark regarding centerline estimations. That's why we state our results for future comparison.

\paragraph{Data augmentation}

To enlarge the datasets and reduce the susceptibility for overfitting, we employ several data augmentation techniques during training.
For all datasets, random color jitter on brightness ($\pm50\%$), contrast ($\pm50\%$), saturation ($\pm50\%$) and hue ($\pm20\%$), random erasing (\SIrange[range-phrase = --]{2}{15}{\percent}) and normalization of the colors within the datasets are applied.
A small random rotation of up to \SI{36}{\degree} and random crop up to \SI{18}{\percent} of the image size is applied to the TuSimple and Argoverse datasets, while full $360^{\circ}$ rotation, but the same cropping are used for the KAI dataset.
\section{Experiments}
\label{sec:experiments}

We evaluate three major parameter choices on the TuSimple dataset, but also show example results for the detection on Argoverse and \acl{KAI} later on.
At first, we examine the influence of the line representation.
Secondly, we investigate how different numbers of predictors for each spatial cell affect the performance. 
And, lastly, we train networks for the different available grid resolutions (\SI{32}{px}, \SI{16}{px}, \SI{8}{px}) and compare the results.

\begin{figure}[t]
	\centering

	\begin{subfigure}[t]{0.24\linewidth}
		\includegraphics[width=1\linewidth]{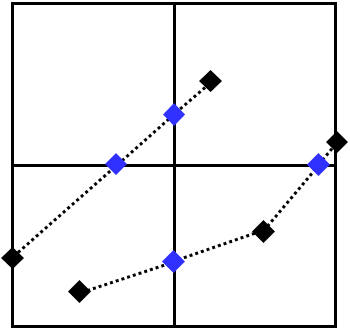}
		\caption{slice}
		\label{fig:slice}
	\end{subfigure}	
	\begin{subfigure}[t]{0.24\linewidth}
		\includegraphics[width=1\linewidth]{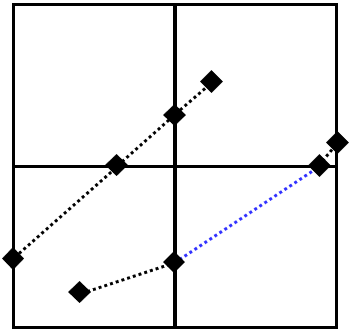}
		\caption{merge}
		\label{fig:merge}
	\end{subfigure}	
	\begin{subfigure}[t]{0.24\linewidth}
		\includegraphics[width=1\linewidth]{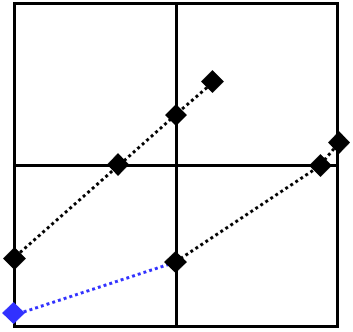}
		\caption{extrapolate}
		\label{fig:extrapolate}
	\end{subfigure}	
		\begin{subfigure}[t]{0.24\linewidth}
		\includegraphics[width=1\linewidth]{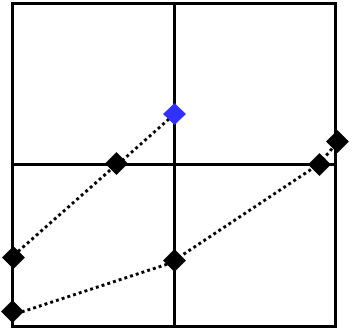}
		\caption{drop}
		\label{fig:drop}
	\end{subfigure}	
	\caption{Converting original polylines to discretized cells on a $2x2$ grid. Black lines visualize the ground truth lines, that are modified from steps a to d. The modifications are highlighted in blue.}
\label{fig:preprocess}
\end{figure}

For comparison, we trained all TuSimple networks for 80~epochs with a batch size of 32 on a Nvidia GTX 1080Ti GPU. The initial learning rate is set to $10^{-3}$. For \acf{lreppo} a learning rate of $ 10^{-4}$ was necessary for convergence. 
Further details on the experiment can be found in the supplementary material.
\paragraph{Pre-processing}
Both \acl{lrep1d} and \acl{lrepeu} force all line segments to start and end on the border of a single grid cell.
Naturally, bounded polylines do not always follow this restriction.
Consequently, we need to convert arbitrarily starting and ending polylines into line segments that are restricted to start and end on cell borders (see \Cref{fig:preprocess}).
First, we calculate the intersection points of the polylines with the cell grid (see \Cref{fig:slice}) and remove points within the cells (see \Cref{fig:merge}). Next, the start and end points of the polylines are extrapolated (see \Cref{fig:extrapolate}) if they cover more than half of the grid cell. Otherwise they are dropped (see \Cref{fig:drop}).

This 
introduces slight discretization errors
which 
are most noticeable in sharp turns and on the ends of a polyline.
Our evaluation accounts for these errors as we compare the predictions to the raw ground truth lines.
However, for clarity, we provide the average ground truth deviation for the different datasets and grid resolutions in the supplementary material.

\paragraph{Line segment evaluation metrics}

To enhance comparability between the different problem settings, we also develop an evaluation metric that allows us to quantify the correctness of a collection of line segments with respect to a set of ground truth polylines. We provide recall, precision and F1 score for all experiments in order to provide an insight into the quality of the actual predictions. 
Other works resort to a pixel-based IoU metric, which is often more difficult to interpret and neglects the direction of line segments~\cite{lsnet}.

For evaluation, we run our \ac{NMS} and sample evaluation points $\hat{p}=(x,y)$ along the ground truth and the predicted line segments (with a sample distance of \SI{1}{px}).
For all points the orientation $\alpha$ of the line segment is calculated in order to compare also the orientation.

Having this representation, we treat predicted points as true positive whenever they are the closest prediction to a ground truth point and lie within a certain radius $\tau$ of all three dimensions $(x,y,\alpha)$.

\paragraph{TuSimple benchmark}

In order to compare our results to related work, we also present the TuSimple benchmark metrics with average accuracy, false negative and false positive rates. 
The TuSimple benchmark provides evaluation scripts and labeled test data for the TuSimple dataset. 
Here, lane boundary instances are expected and thus further post-processing is required.
First, we propose to 
find connected components
in the generic \ac{NMS} output. At each level we calculate the average representative and then fit a cubic spline into each line instance.

We assume two \ac{NMS} predictions $a$ and $b$ to be connected if the start point of $a$ is within a certain distance from $b$'s endpoint (here: \SI{75}{\percent} of the cell size). This is done using a breadth-first search with a loop detection, resulting in a distinct tree-like structure for each component. Next, for each tree, predictions are averaged across the same depth level. Finally, to increase fidelity, each connected component is interpolated with a cubic spline.
Further details about the TuSimple post-processing can be found in the supplementary material.
\begin{table}[tb]
	\centering
	\begin{tabular}{|l|lll|lll|}
		\hline
		Line 		 & Acc 			    & FP 			& FN 			 & F1 		    & Rec. 	    & Prec.         \\ \hline \hline
		\acs{lrep1d} & \snd{.887} 		& \snd{.157}    & \snd{.160}	     & .616         & \snd{.955}    & .455          \\ %
		\acs{lrepeu} & .877             & \first{.137}  & .168           & \snd{.685}   & \first{.956}  & \snd{.533}    \\ %
		\acs{lreppo} & \first{.914}     & .159          & \first{.112}   & \first{.740} & .950          & \first{.607} \\ %
		\hline
	\end{tabular}
	\caption{Results for different line representations on the TuSimple validation set. We evaluate \acl{lrep1d}, \acl{lrepeu} and \acl{lreppo}. The architecture has eight predictors per \SI{32}{px} cell.}
	\label{tab:eval_linerep}
\end{table}

\paragraph{Line representations}
One of the major modeling decisions of our approach lies in the choice of a suitable line representation.
It is worth mentioning that the \acf{lrep1d} converge rather slowly compared to \acf{lrepeu} due to the fact that the network has to overcome the discontinuity between points before and after the top left corner of the grid cell, where the mathematical distance is huge, but the actual difference is tiny. Further, in order to achieve convergence with \acf{lreppo}, it appears to be useful to initialize a smaller learning rate compared to the other two experiments.

We present three experiments comparing the representations in \Cref{tab:eval_linerep}.
For each, we used eight predictors per cell, a cell size of \SI{32}{px} (i.e. no upsampling).
Unbound Cartesian Points (Po) have the highest false positive rate for TuSimple, but are the most convincing regarding both generic F1 score and TuSimple accuracy.
We assume this to be due to the flexibility that comes with this representation.
While we liked the idea of the minimal representation of \acl{lrep1d}, it proved to be unstable during training.
Hence, we will perform further experiments only on \acl{lreppo} and \acl{lrepeu}.

\begin{table}[tb]
	\centering
	\begin{tabular}{|l|lll|lll|}
		\hline
		Line Pred & Acc 			& FP 			& FN 			& F1 		& Rec. 			& Prec.          \\ \hline \hline
		\acs{lrepeu}$\>$ $\>$ $\>$ $\>$4  	& \snd{.878}    & .168          & .180          & .546          & \first{.971}      & .380  	    \\ %
		\acs{lrepeu}$\>$ $\>$ $\>$ $\>$8  	& .877          & \snd{.137}    & \snd{.168}    & \snd{.685}    & \snd{.956}        & \snd{.533}  	\\ %
		\acs{lrepeu}$\>$ $\>$ $\>$ 12  	& \first{.885}  & \snd{.154}    & \first{.159}  & \first{.717}  & .940              & \first{.580}   \\  %
		\hline
		\acs{lreppo}$\>$ $\>$ $\>$ $\>$4  	& \first{.922}  & \snd{.157}    & \first{.099}  & .713          & \first{.952}      & .571 			\\ %
		\acs{lreppo}$\>$ $\>$ $\>$ $\>$8  	& \snd{.914}    & .159          & \snd{.112}    & \snd{.740}    & \snd{.948}        & \snd{.607}    \\ %
		\acs{lreppo}$\>$ $\>$ $\>$ 12  	& .903          & \first{.135}  & .120          & \first{.777}  & .941              & \first{.662}  \\ %
		\hline
	\end{tabular}
	\caption{Results on different number of predictors for both \acf{lreppo} and \acf{lrepeu} on the TuSimple validation set without any upsampling (\SI{32}{px}).}
	\label{tab:eval_preds}
\end{table}

\paragraph{Number of predictors}

Another important aspect in designing single shot detection models is the allowed number of
predictions per cell.
Therefore, we trained networks with 4, 8 and 12 predictors per cell at a default resolution of $32\times32$\,pixels. The results are presented in \Cref{tab:eval_preds}. They suggest that fewer predictors work better for TuSimple accuracy while more predictors work better for the more generic F1 score. Hence, we recommend 12 predictors for more complex applications while using four or eight predictors for the TuSimple lane boundary estimation task.

\paragraph{Grid resolution}
The third important choice is whether to use the default resolution of $32\times32$\,pixels or add one ($16\times16$\,px) or two upsampling modules ($8\times8$\,px). 
We evaluate all grid resolutions on both \acf{lrepeu} and \acf{lreppo} with eight predictors per cell (see \autoref{tab:eval_grid}). 
Using one upsampling module (\SI{16}{px}) seems to be best for TuSimple accuracy while two upsampling modules (\SI{8}{px}) is the winner regarding generic F1 score.

\begin{table}[tb]
	\centering
	\begin{tabular}{|l|lll|lll|}
		\hline
		Line Grid           & Acc 		    & FP 	        & FN            & F1            & Rec. 	    & Prec.         \\ \hline \hline
		\acs{lrepeu}$\>$ $\>$ 32\,px & \snd{.877}    & \first{.137}  & \first{.168}  & \snd{.685}    & \snd{.956}    & \snd{.533}    \\ %
		\acs{lrepeu}$\>$ $\>$ 16\,px & \first{.899}  & \snd{.343}    & \snd{.201}   	& .658       	& \first{.959} 	& .500          \\ %
		\acs{lrepeu}$\>$ $\>$ 8\,px  & .839          & .481          & .331  	    & \first{.712}  & .919          & \first{.581}  \\  %
		\hline
		\acs{lreppo}$\>$ $\>$ 32\,px & \snd{.914}    & \first{.159}  & \snd{.112}    & \snd{.740}     & \snd{.948}    & \snd{.607}    \\ %
		\acs{lreppo}$\>$ $\>$ 16\,px & \first{.930}  & \snd{.258}    & \first{.095}  & .739          & \first{.951}  & .604          \\ %
		\acs{lreppo}$\>$ $\>$ 8\,px  & .875          & .405          & .196          & \first{.776}  & .930          & \first{.665}  \\ %
		\hline
	\end{tabular}
	\caption{Results for different cell sizes on the TuSimple validation set. We compare \acl{lrepeu} and \acl{lreppo} with eight predictors. \SI{16}{px} is best in TuSimple accuracy, \SI{8}{px} has the highest generic F1 score.}
	\label{tab:eval_grid}
\end{table}

\paragraph{Comparison with the state of the art}

As there are no benchmarks for semantic polyline detection, we settled for the TuSimple lane boundary estimation benchmark as simple example application. Thus, our generic framework competes with approaches that are tailored, but also limited to the very detection of a few mostly vertical line instances.

For evaluation on the test set, the previous experiments suggested two candidate configurations: \acl{lreppo} at a resolution of \SI{16}{px} with either four or eight predictors. Since, on the validation set, four predictors only scored 88\% accuracy, we discarded this option. 

We did not expect to win, but rather wanted to evaluate performance w.r.t. far more specialized networks. With 94.2\% accuracy on the test set (see \autoref{tab:winner}), our most promising candidate gets within three percentage points of the best specialized competitor, which is a lot better than we expected. In \Cref{fig:tus_results}, we show sample results.

Hence, while not setting a new state of the art for highway lane boundary detections, YOLinO is so far the only choice for many, especially more complex applications which 
do not have a multitude of existing, specialized solutions. 
Still, we get close to the best possible performance without requiring any network specialization.
\begin{table}[tb]
	\centering
	\begin{tabular}{|l|lll|r|}
		\hline
		Method                  & Acc           & FP        & FN        & fps       \\ \hline \hline
		LineCNN~\cite{Li2020LineCNN}&\first{.969}& .044     & \snd{.020}& 17	    \\	
		PINet~\cite{ko2020key}  & .958          & .059      & .033      & 40        \\
		LaneATT~\cite{tabelini2020keep}&  \snd{.967}& \snd{.036}  	& \first{.018}& \snd{250}	\\	
		ResNet-18~\cite{qin2020ultrafast} &  .961 & \first{.019}\footnote{reported by~\cite{tabelini2020keep}}
		& .040\footnotemark[2]		    & \first{312}	    \\	
		
		\hline
		\textbf{YOLinO (ours)} & .942          & .188      & .076      & 187       \\ %
		\hline 
	\end{tabular}
	\caption{Our best-performing configuration (\acs{lreppo}, eight predictors, \SI{16}{px}) compared to selected related work that are specifically tailored towards lane boundary detection. While YOLinO is not tailored to this task, we are able to reach comparable accuracy on the TuSimple test set. Note that regarding inference speed, YOLinO is among the fastest approaches.}
	\label{tab:winner}
\end{table}

\paragraph{Other applications}

To show the wide range of applications for our approach, we present further results on \acl{KAI} and Argoverse. 

The Argoverse dataset is by far the most challenging task in the evaluation. Lane centerlines are already difficult for humans to determine and, thus, labels are also prone to variations. The detected centerlines have a precision of \SI{41}{\percent}, recall of \SI{69}{\percent} leading to an F1 score of \SI{52}{\percent}. Examples can be found in~\Cref{fig:argo_results}. Regarding the complexity of lane structures, this is already a great result. 

\begin{figure*}
    \centering
	\begin{subfigure}{0.97\textwidth}
        \makebox[20pt]{\raisebox{40pt}{\rotatebox[origin=c]{90}{\footnotesize{Ground truth}}}}%
		\centering
		\includegraphics[width=0.315\textwidth,trim=0 100 0 100, clip]{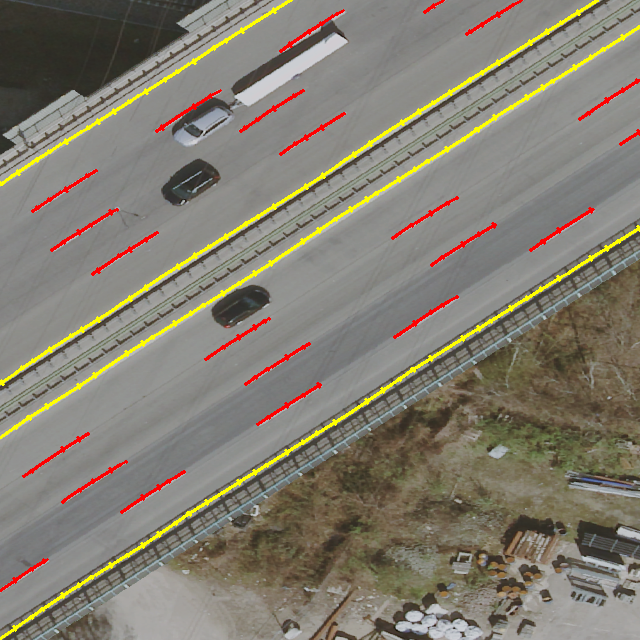} 
		\includegraphics[width=0.315\textwidth,trim=0 100 0 100, clip]{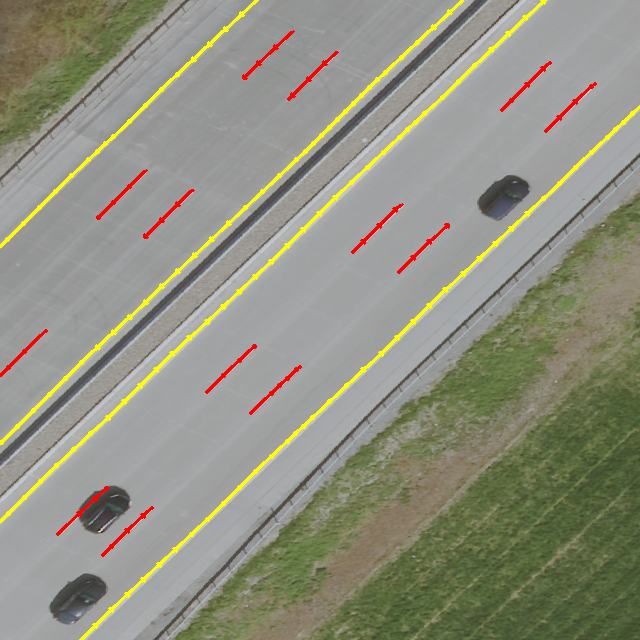}
		\includegraphics[width=0.315\textwidth,trim=0 100 0 100, clip]{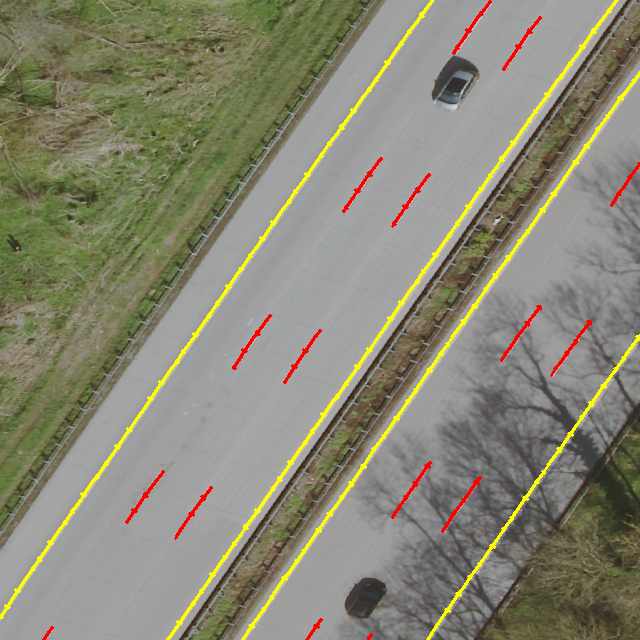}
	\end{subfigure}
	\begin{subfigure}{0.97\textwidth}
		\centering
        \makebox[20pt]{\raisebox{40pt}{\rotatebox[origin=c]{90}{\footnotesize{Prediction after NMS}}}}%
		\includegraphics[width=0.315\textwidth,trim=0 100 0 100, clip]{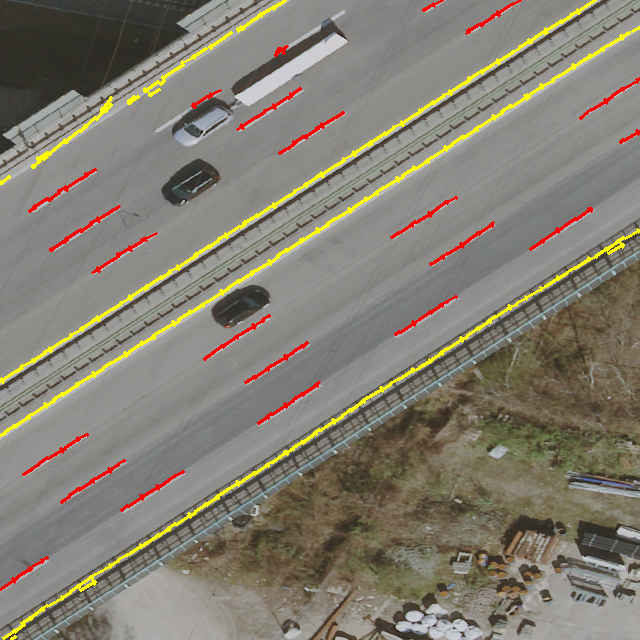} 
		\includegraphics[width=0.315\textwidth,trim=0 100 0 100, clip]{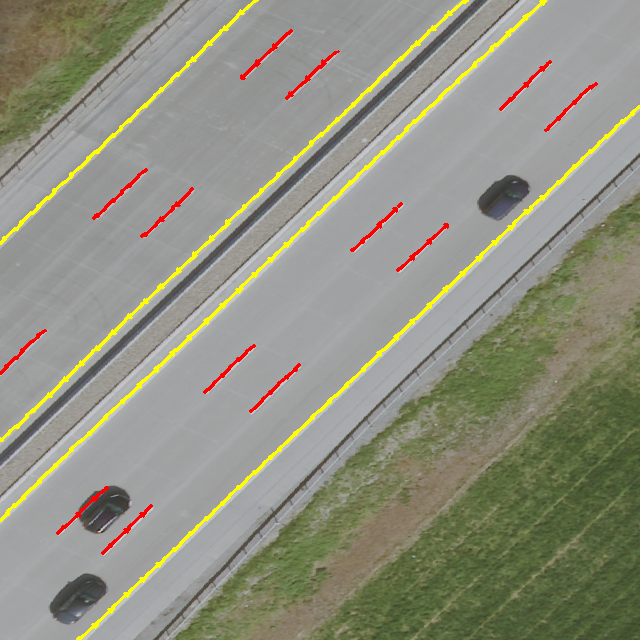}
		\includegraphics[width=0.315\textwidth,trim=0 100 0 100, clip]{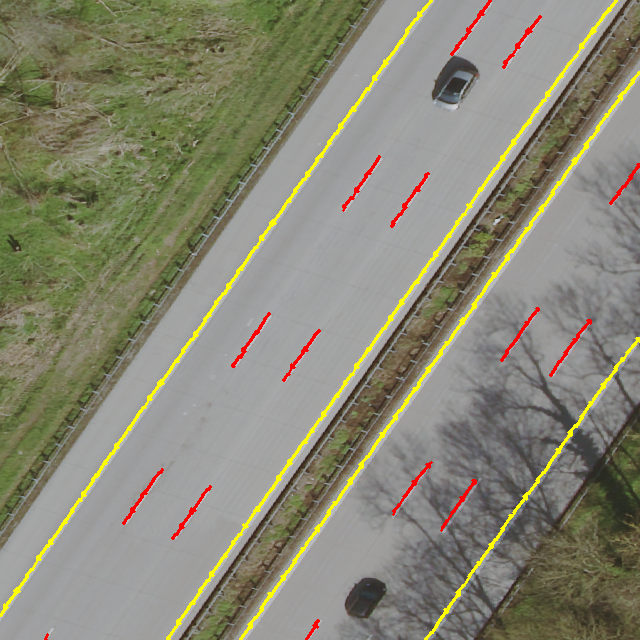}
	\end{subfigure}
	\caption{Results for the \acs{KAI} dataset. The network has eight predictors, \SI{16}{px} cells and \acl{lreppo}. Colors indicate the different classes: dashed and solid road marking. Aerial images: \textcopyright~City of Karlsruhe $\vert$ Liegenschaftsamt}
	\label{fig:aerial_results}
\end{figure*}

\acl{KAI} contain mostly homogeneous images, but only few examples of corner cases like on-ramps or bridges.
Still, we are able to cope with these challenges and predict markings with an F1 score of \SI{89}{\percent}, a recall of \SI{90}{\percent} and a precision of \SI{88}{\percent}. 
Examples can be found in \autoref{fig:aerial_results}.

\section{Conclusion}

We presented \textit{YOLinO}, a single shot polyline detector enabling to build real-time applications using polylines.
While being a generic approach, YOLinO is comparable to state-of-the-art TuSimple lane detection algorithms.
Besides being extremely fast, our approach can detect a wide range of polylines, even if they are not explicitly visible, branching or even crossing. 
Moreover, both dashed and continuous lines can be detected with the same head.
This enables robotic applications such as road marking detection or lane centerline estimation, 
but is also conceivable for many others, \eg blood vessel detection. 
While we demonstrated the general idea using YOLO9000, using a modern backend such as EfficientNet~\cite{tan2019efficientnet} and incorporating ideas from YOLOv4~\cite{YOLOv4} is expected to improve performance even more. 
This also holds for adding a learned \acs{NMS} module, e.g. exploiting the grid structure with a Graph Neural Network.

\clearpage

\begin{figure*}[h]
    \centering	
	\begin{subfigure}{0.32\linewidth}
		\centering
 		\includegraphics[width=1\linewidth,trim=35 200 600 200, clip]{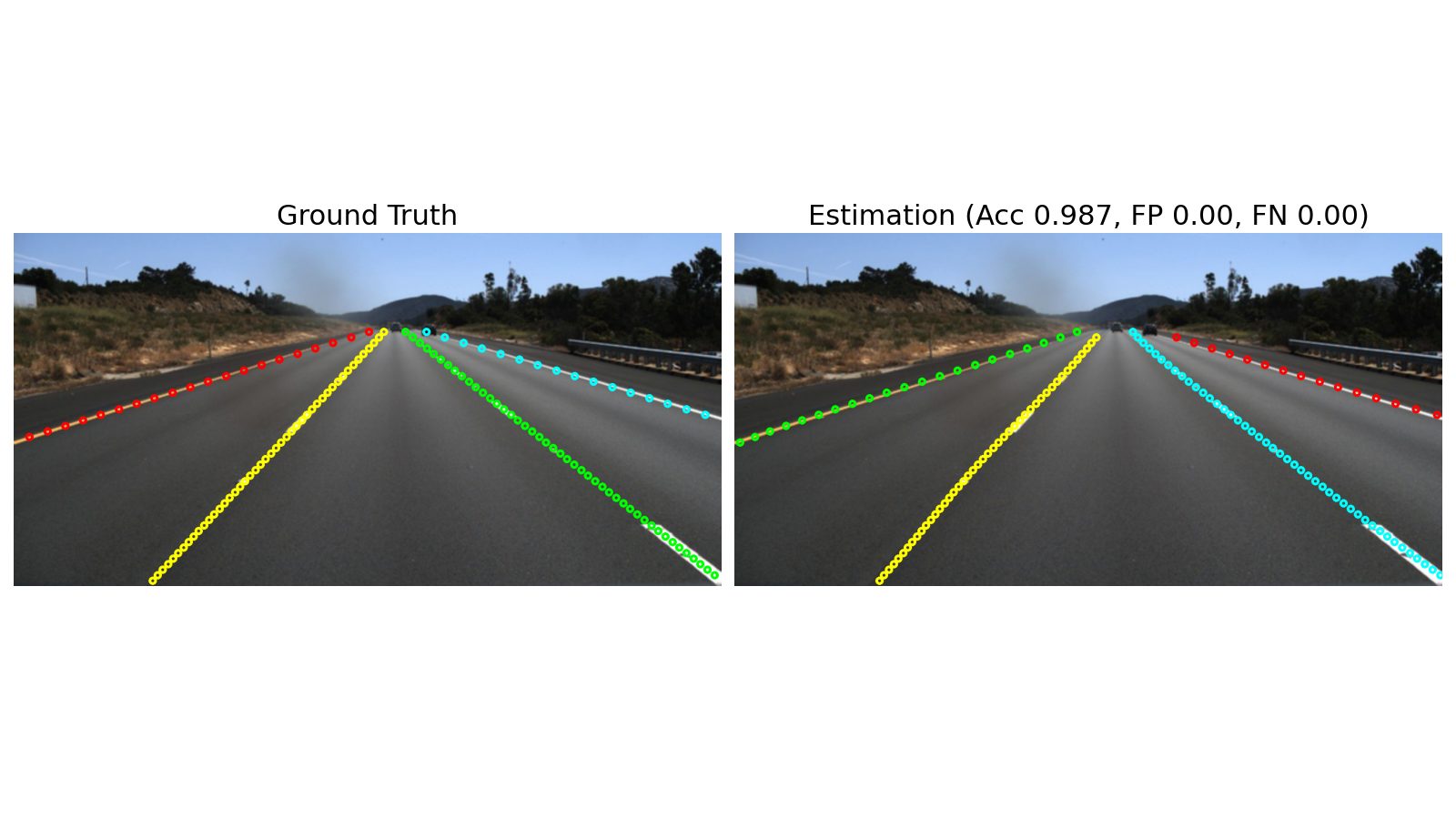} %
		\centering
		\includegraphics[width=1\linewidth,trim=35 200 600 200, clip]{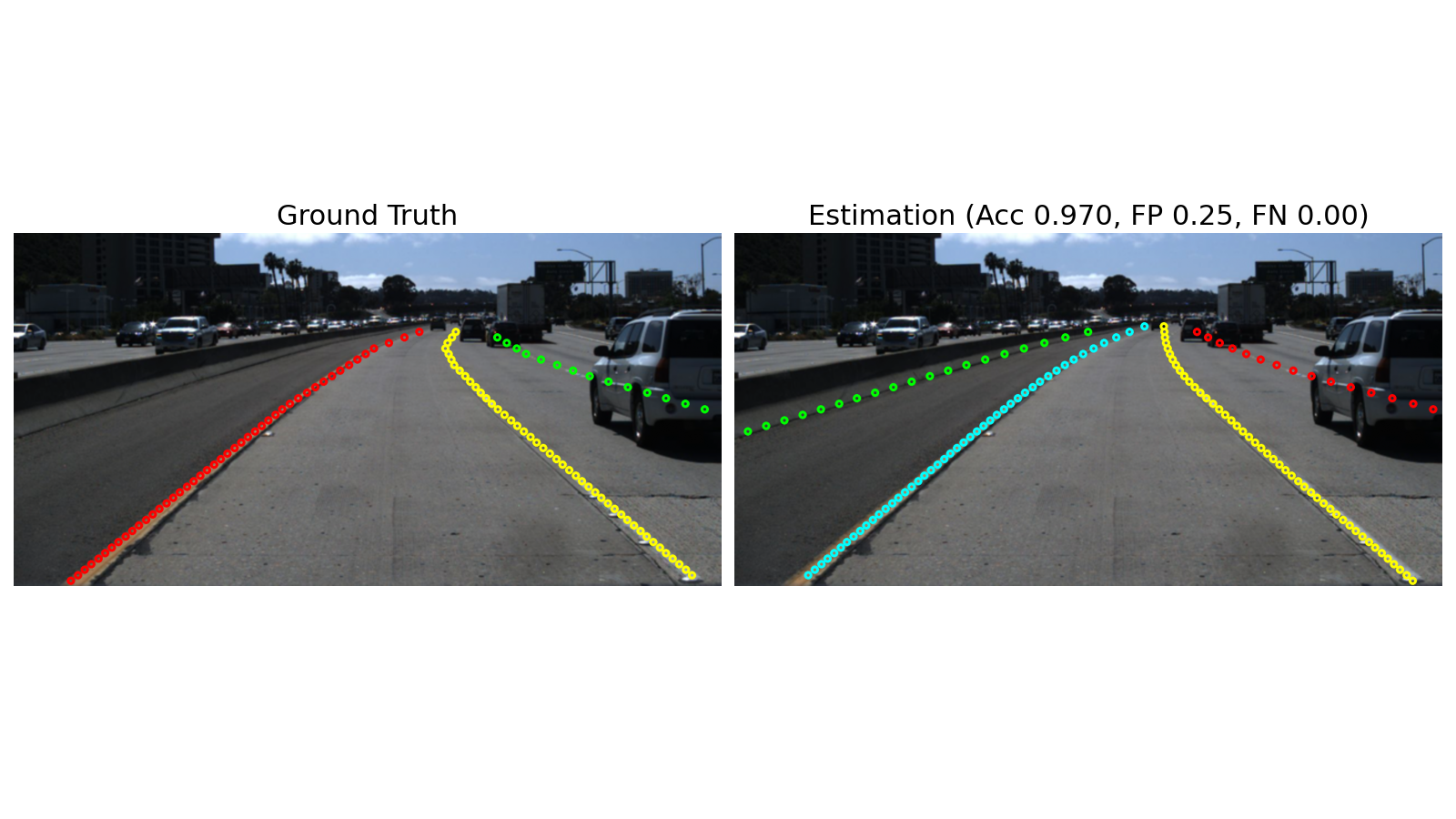} %
		\centering
		\includegraphics[width=1\linewidth,trim=35 200 600 200, clip]{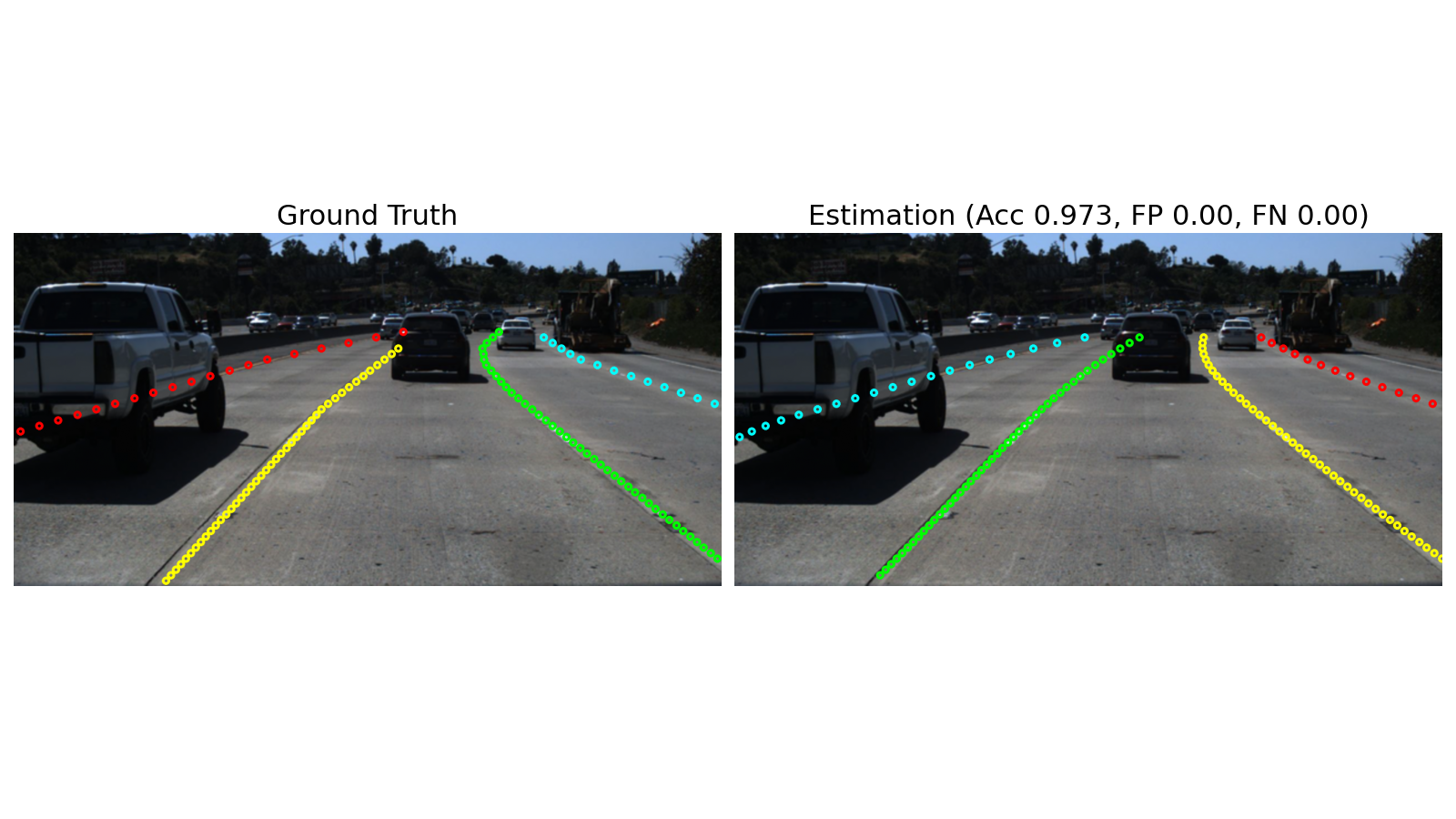} %
		\caption{Ground truth}
	\end{subfigure}
	\begin{subfigure}{0.32\linewidth}
	    \centering
        \includegraphics[width=1\linewidth,trim=30 18 25 20, clip]{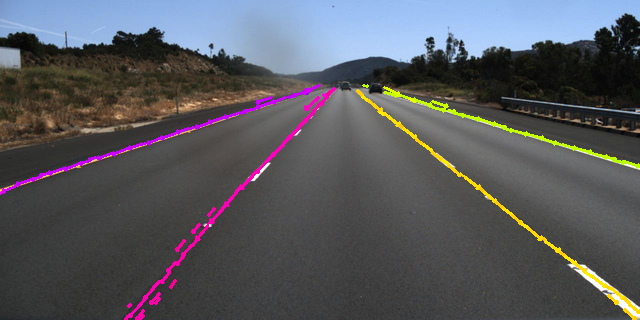}
        \centering
        \includegraphics[width=1\linewidth,trim=30 18 25 20, clip]{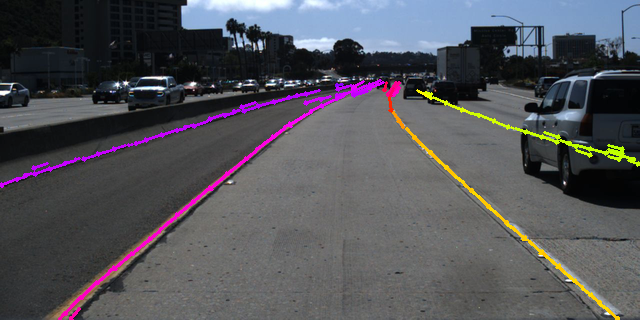}
        \centering
        \includegraphics[width=1\linewidth,trim=30 18 25 20, clip]{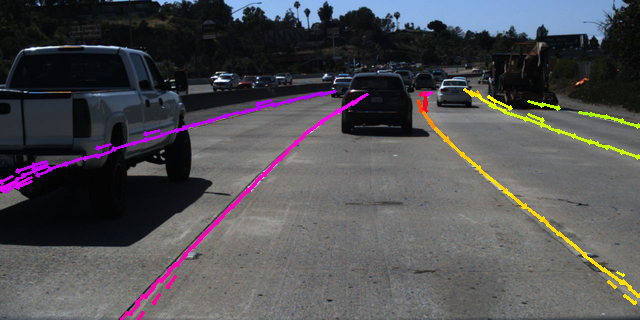}
		\caption{Prediction}
	\end{subfigure}
	\begin{subfigure}{0.32\linewidth}
	    \centering
        \includegraphics[width=1\linewidth,trim=600 200 35 200, clip]{res/\tus/tusimple/0530_1492626875719975670_0.png}
        \centering
        \includegraphics[width=1\linewidth,trim=600 200 35 200, clip]{res/\tus/tusimple/0530_1492638943713602995_0.png}
        \centering
        \includegraphics[width=1\linewidth,trim=600 200 35 200, clip]{res/\tus/tusimple/0531_1492638617945022881.png}
		\caption{Prediction post-processed}
	\end{subfigure}
	\caption{Results for the TuSimple benchmark using \acl{lreppo}, eight predictors and a grid resolution of \SI{16}{px}. Colors in (a) and (c) indicate instances, but do not need to be similar for prediction and ground truth. In (b) the colors visualize the orientation of the predicted line segments.}
	\label{fig:tus_results}
\end{figure*}
\begin{figure*}[tb]
    \centering
	\begin{subfigure}{0.32\linewidth}
		\includegraphics[width=1\linewidth]{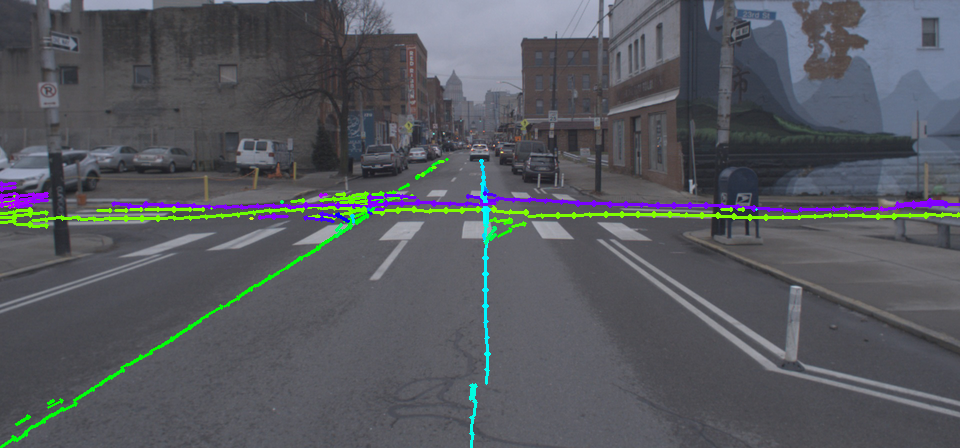} 
	\end{subfigure}
	\begin{subfigure}{0.32\linewidth}		
		\includegraphics[width=1\linewidth]{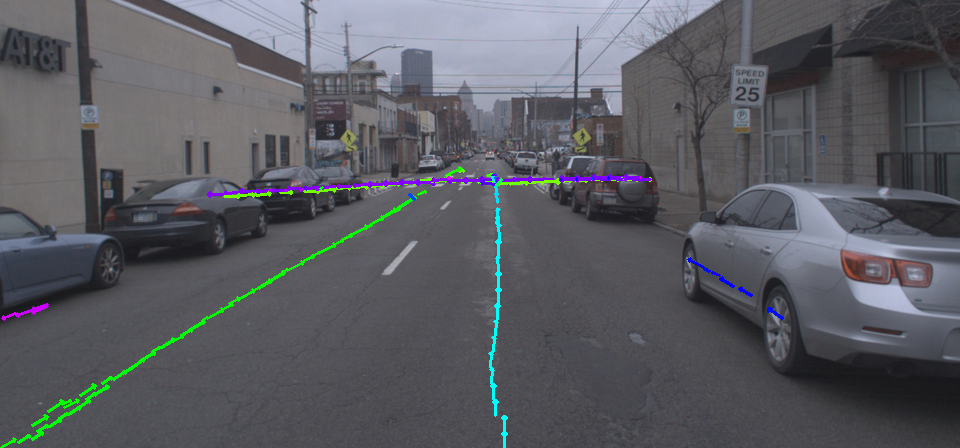} 		
	\end{subfigure}
	\begin{subfigure}{0.32\linewidth}		
		\includegraphics[width=1\linewidth]{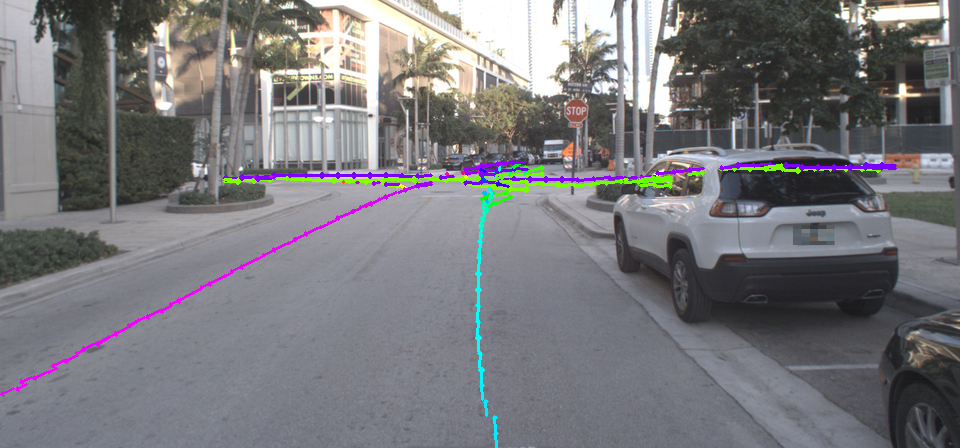} 		
	\end{subfigure}
	\caption{Results for the Argoverse lane centerline task. Colors encode the the driving direction of a lane. 
	In the first two images, YOLinO is able to correctly detect a one-way street with both lanes driving in the same direction. In all three images, detecting intersections does not pose a problem as well.
	}
	\label{fig:argo_results}
\end{figure*}

\clearpage
\clearpage
\clearpage
\appendix
\section*{Appendix}
In this supplementary material, we provide further details on YOLinO. In \Cref{sec:architecture}, the base architecture is explained in detail. Here, we also provide training parameters. 
An estimation of the discretization errors can be found in \Cref{sec:discretization}. 
For further detail on the \ac{NMS} or the TuSimple-specific post-processing, please refer to \Cref{sec:post_proc}. 
Following, we describe the calculation of our evaluation metrics in \Cref{sec:scores}, followed by an overview of all experiments of the ablation study in \Cref{sec:results}. Qualitative results from all three datasets can be found in \Cref{sec:tus}, \Cref{sec:aer} and \Cref{sec:argo}. 
\section{Architecture}
\label{sec:architecture}

\begin{figure*}[tb]
    \centering
    \includegraphics[width=0.8\textwidth]{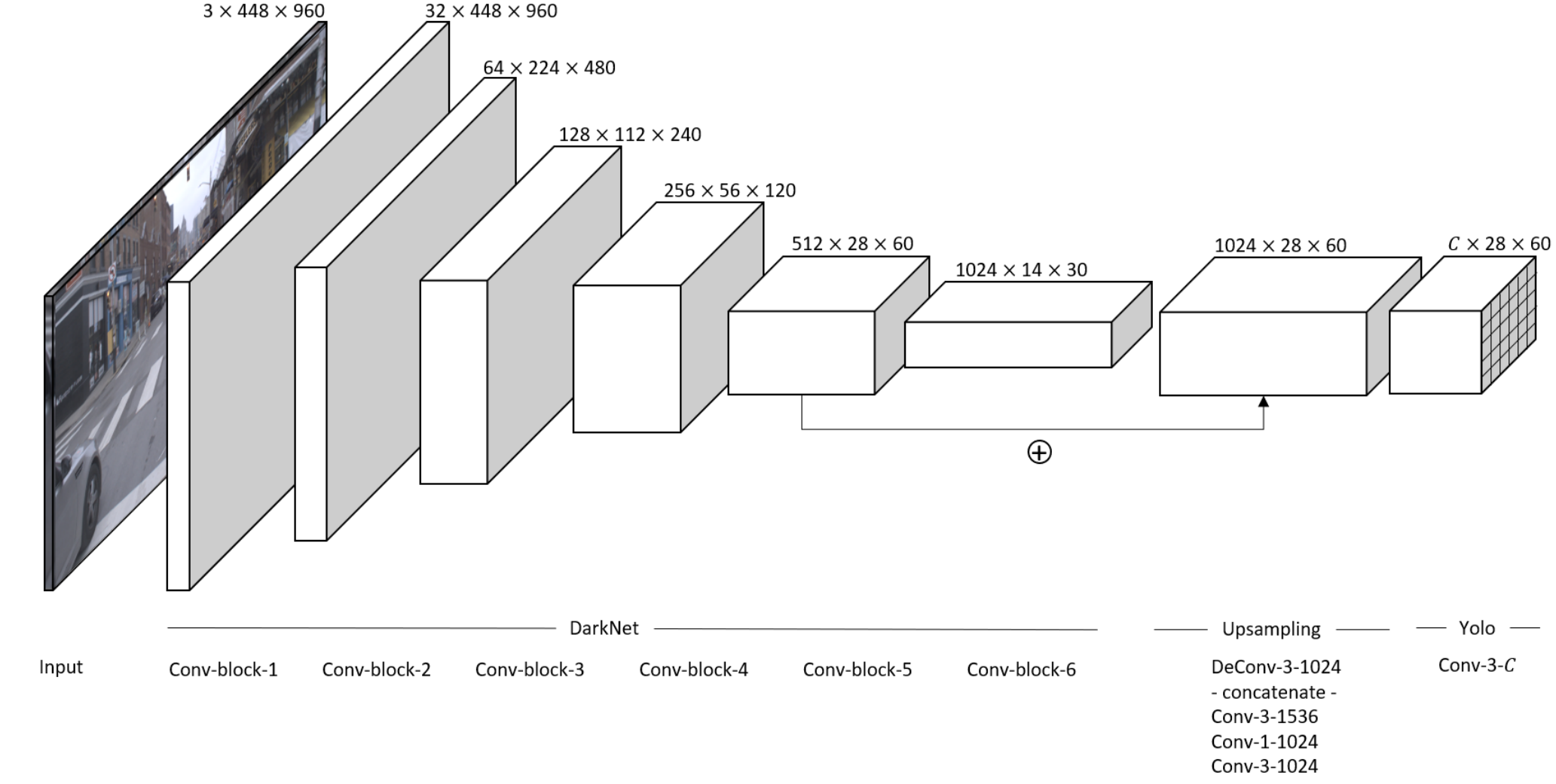}
    \caption{YOLinO's feed-forward architecture. Here shown for the Argoverse input images and with a single upsampling stage, leading to grid cells of $16\times16$\,\si{px}. Each grid cell predicts up to 12 line segments with a geometric definition $g$, a confidence value $c$ and an optional classification $l$, all together resulting in the output channels $C$.}
    \label{fig:architecture}
\end{figure*}

For the backbone we chose YOLO9000 with Darknet-19~\cite{yolov2}. Depending on the required resolution of the output grid, we employ zero, one or two upsampling blocks. In order to pass local features, we further implement a skip connection between the downsampling and upsampling layers. \Cref{fig:architecture} shows the full architecture with one upsampling block, resulting in cells of $16\times16$\,\si{px}. 

The training is implemented in Pytorch~\cite{Ketkar2017pytorch} and parameterized according to the configuration in \autoref{tab:train_config}.

\begin{table}[b]
    \centering
    \begin{tabular}{|l|lll|}
    \hline              & Argoverse                & \acs{KAI}                  & TuSimple                 \\
    \hline \hline
    {Batch Size}              & {16}                     & {16}                       & {32}                      \\

    {Input Image Size}         & 448x960                  & 640x640                    & 320x640                  \\
    {Decay Rate}               & {$10^{-4}$}              & {$10^{-4}$}                & {$10^{-4}$}                  \\
    {Optimizer}                & {Adam}                   & {Adam}                     & {Adam}                    \\
    Epochs                    &  55                     &  30                        &  40                        \\

    \hline
    \end{tabular}
    \caption{List of training parameters of each experiment shown in qualitative results. Epochs until convergence are for \acl{lreppo} with eight predictors and \SI{16}{px} resolution.}
    \label{tab:train_config}
\end{table} %

\section{Discretization}
\label{sec:discretization}

Especially for \acl{lrep1d} and \acl{lrepeu} binding start and endpoint to the cell borders introduces slight errors.
These deviations are most noticeable in sharp turns and on the ends of a polyline. Our recall already accounts for these errors as we compare the predictions to the full ground truth lines. However, for clarity we provide the average ground truth deviation for the different datasets and grid resolutions in \Cref{tab:slicing_errors}. 
As expected, the introduced error is smaller the higher the grid resolution.
We also see higher values for the \ac{KAI} dataset as it contains many short polylines that do not initially align with the grid. %
Similarly, the Argoverse dataset features many splitting polylines and contains shorter polylines in the more distant road areas.
Contrastingly, the polylines in TuSimple are mostly continuous which makes the overall deviation less distinctive.
We conclude that the most suitable grid resolution depends on the specific problem setting as finer grids also introduce significantly higher computational costs.

\begin{table}[tb]
\centering
\begin{tabular}{|cccc|}
\hline
Size (px) & TuSimple  & KAI & Argoverse \\ \hline \hline
32$\times$32      & 1.40       & 1.85 & 2.57      \\
16$\times$16      & 0.42       & 0.55 & 0.79      \\
8$\times$8       & 0.14       & 0.16 & 0.25      \\ \hline
Input Size (px) & $320\times640$  & $640\times640$ & $448\times960$ \\ \hline
\end{tabular}
\caption{Ground truth deviation induced by pre-processing with respect to different grid resolutions, given as average pixel difference for all line segments.}
\label{tab:slicing_errors}
\end{table}

\begin{figure*}[tb]
	\begin{center}	
		\begin{subfigure}{0.49\linewidth}
			\centering
			\includegraphics[width=1\linewidth]{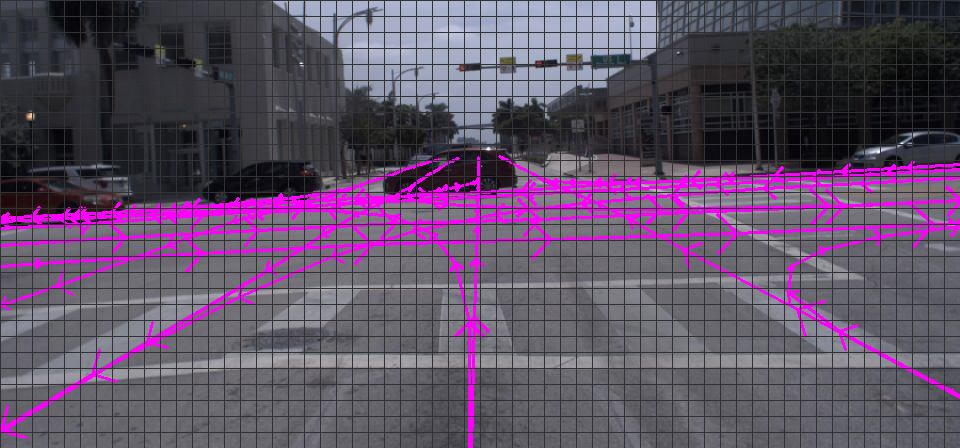}
			\caption{Original Argoverse ground truth}
		\end{subfigure}
		\hfill
		\begin{subfigure}{0.49\linewidth}
			\centering
			\includegraphics[width=1\linewidth]{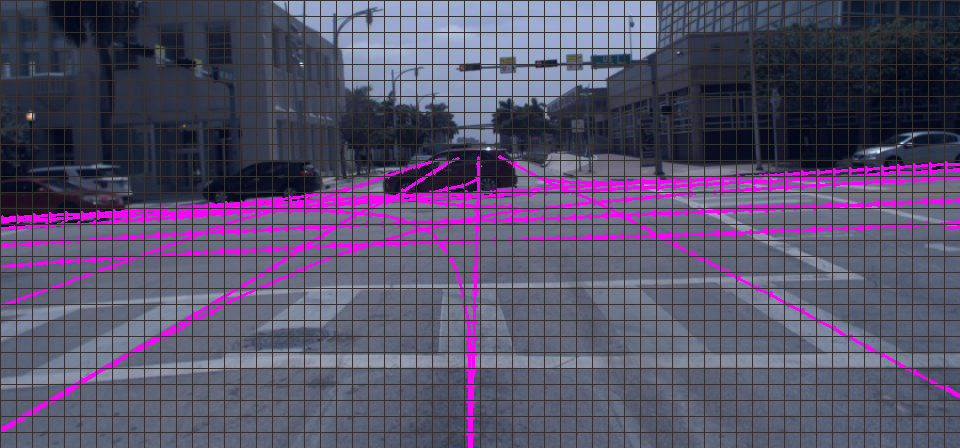}
			\caption{Discretized ground truth}
		\end{subfigure}
	\caption{Comparison of the original Argoverse ground truth center lines with the discretized grid lines used for training.}
	\end{center}
\end{figure*}

\section{Post-processing}
\label{sec:post_proc}
As our representation in terms of raw predictors is generic, but the approach is applicable to many concrete applications, we distinguish two kinds of post-processing. First, we describe a generic \acf{NMS} that is beneficial for almost any application since it suppresses redundant predictors. Second, we show how our generic output can be post-processed such that the output is comparable with that of neural networks specialized to these very applications. For this, we chose the TuSimple lane boundary estimation benchmark.

\paragraph{Generic \ac{NMS}}

For the generic NMS, we expect predictors $P = (g, l, c)$. First, we discard all predictors with $c \leq \tau_c$ (cf. \Cref{tab:nms_parameters}). Next, we convert the various geometric representations of a predictor to a common one that is more suitable for DBSCAN, hereafter called \emph{\ac{NMS} coordinates} $\tilde{g}$. Here, $\tilde{g} = \left(m_x, m_y, \ell, d_x, d_y\right)^\intercal$ consists of the midpoint of the line segments in image coordinates $m_x, m_y$, allowing to cluster even across neighboring cells. $\ell$ is the length of the predictors in image coordinates. Finally, $d_x, d_y$ are the \emph{normalized} directions for each predictor. For instance, for the \acf{lreppo} representation, we have $g_{\text{Po}} = (g_s, g_e)$ with $g_s,g_e \in \left[0, 1\right]\times\left[0, 1\right]$. We first convert $g_e, g_s$ to image coordinates, yielding $\hat{g}_s, \hat{g}_e \in \left[0, M\right]\times\left[0, N\right]$. These can then be used to calculate the common \ac{NMS} coordinates:

\begin{align}
    \begin{pmatrix}
        m_{x} \\
        m_{y}
    \end{pmatrix} &= \lambda_m \kappa \frac{\hat{g}_{e} + \hat{g}_{s}}{2}\\
    \Delta &= \begin{pmatrix}
        \delta_{x} \\
        \delta_{y}
    \end{pmatrix} = \hat{g}_{e} - \hat{g}_{s}\\
    \ell &= \lambda_\ell \norm{\Delta} \\
    d_x &= \lambda_d \frac{\delta_x}{\ell} \\
    d_y &= \lambda_d \frac{\delta_y}{\ell}
\end{align}

The factors $\mathbf{\lambda}$ scale the coordinates appropriately and are determined empirically for each dataset (cf. \Cref{tab:nms_parameters}). Factor $\kappa$ denotes the grid scale,
i.e. $\kappa = 1.0$ for our raw output with $32\times32$\,pixels cell size, $\kappa = 0.5$ for $16\times16$\,pixels etc.

\begin{table*}[tb]
    \centering
    \begin{tabular}{|ll|lll|}
    \hline                  & & Argoverse                & \acs{KAI}                  & TuSimple                 \\
    \hline \hline
    {Confidence threshold} & {$\tau_c$}     & {0.99}                    & {0.95}                      & {0.9}                     \\

    {Weight for segment length} & {$\lambda_\ell$}& {0.016}                  & {0.013}                    & {0.013}                   \\
    {Weight for midpoint position} & {$\lambda_m$}      & {1.5}                      & {1.5}                        & {2}                       \\
    {Weight for directions} & {$\lambda_d$} & {0.05}                   & {0.05}                     & {0.05}                   \\
    \hline
    \end{tabular}
	\caption{List of \acs{NMS} parameters for each dataset.}
	\label{tab:nms_parameters}
\end{table*}

Next, we apply the DBSCAN clustering by Scikit-learn~\cite{Pedregosa2011ScikitLearn} using exponentially scaled confidences as weights $w_i = c_i^{10}$ and DBSCAN parameters $\epsilon = 0.02$ and at least two predictors per cluster (also called \texttt{min\_pts} or \texttt{min\_samples}). Finally, we take the weighted average of each cluster in \ac{NMS} coordinates using the same exponential weights $w_i$ we used for clustering. 
For visualization and downstream applications, we found it more intuitive to take the maximum confidence of each cluster instead of the weighted average.

Both, coordinate conversion and averaging are done in NumPy~\cite{Harris2020Array}. The overall speed of 230~fps holds for all three steps, conversion, DBSCAN and averaging taken together. Pipelining the three steps could even roughly double the throughput with DBSCAN being the bottleneck. 

\paragraph{TuSimple Post-processing}

For TuSimple, in addition to the generic \ac{NMS}, we need to convert predicted line segments into contiguous and smooth, but accurate lane boundary estimates.

First, to prevent cycles, we discard all predictors which point downwards by more than $0.25$, i.e. a quarter of a grid cell. For all remaining predictors, we determine a successor by minimal start/endpoint distance such that the successor start point is closer than $0.75$ grid cells and not in the bottom half of the bottom row of the grid.
\begin{figure}[tb]
    \centering
    \includegraphics{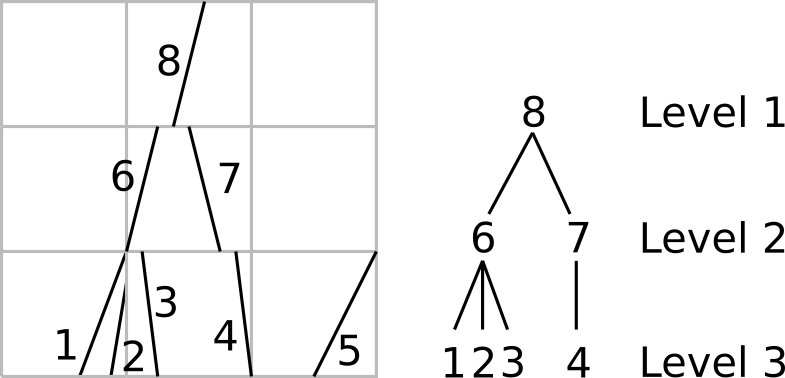}
    \caption{Illustration of the breadth-first search (BFS) and averaging used for polyline extraction, here using $\tau_s = 3$. Segments 1, 2, and 3 share the same successor, as do segments 6 and 7. 8 is a root node and will initiate one BFS procedure leading to the tree on the right with three levels. Each level is averaged, meaning that segments 6 and 7 as well as segments 1-4 are averaged. Segment 5 has no successor in reach and is discarded as the resulting polyline is too short.}
    \label{fig:bfs_tree}
\end{figure}
Given this adjacency and possible roots that have no successor (i.e. usually the topmost predictors in each "tree"), but are possibly successor to multiple predictors, we start a breadth-first search starting from those root nodes (cf.~\Cref{fig:bfs_tree}). At each level, we calculate the weighted average of all predictors at the same level using confidences as weights. This leads to a polyline that is close to the desired output, but not yet smooth. First, we discard polylines with less than $\tau_s = 10$ segments. Second, we fit a B-spline of degree $3$ through the mid points of each predicted line segment using SciPy's~\cite{Virtanen2020SciPy} \texttt{splprep} function with smoothing parameter $s = 0.05$. These splines are then evaluated, leading to the points that we use for evaluation. 

\begin{figure*}[tb]
	\begin{center}	

	\begin{subfigure}{0.4\linewidth}
		\includegraphics[width=1\linewidth,trim=35 200 600 200, clip]{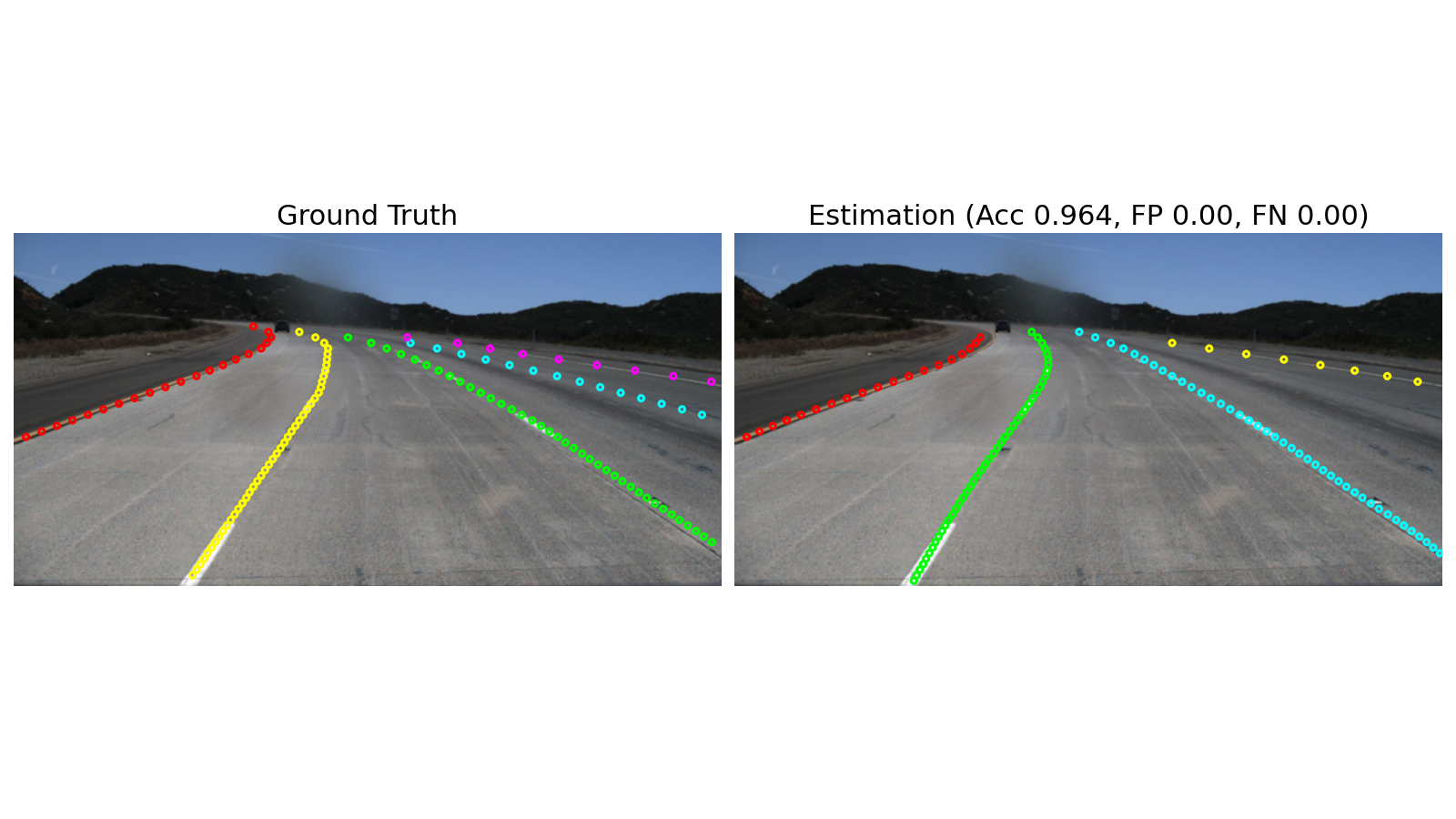}
		\caption{Ground Truth}
	\end{subfigure}
	
	\begin{subfigure}{0.4\linewidth}
		\includegraphics[width=1\linewidth,trim=30 18 25 20, clip]{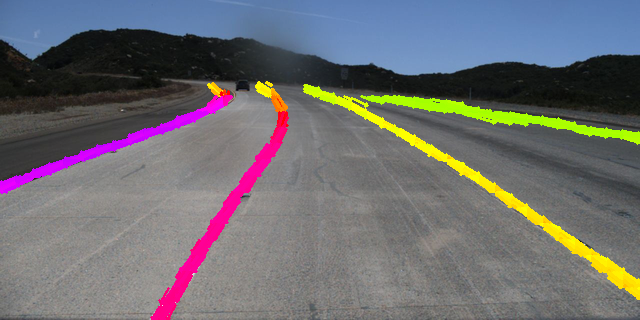}
		\caption{Prediction}
	\end{subfigure}
	\begin{subfigure}{0.4\linewidth}
		\includegraphics[width=1\linewidth,trim=30 18 25 20, clip]{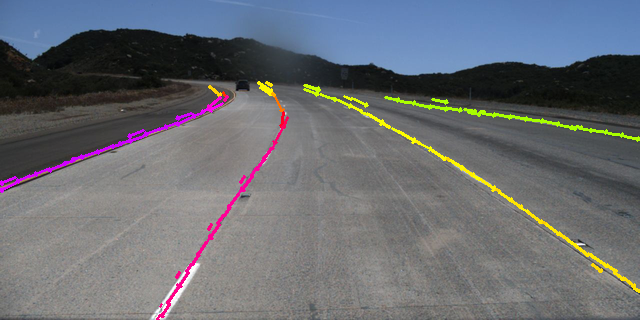}
		\caption{Generic NMS}
	\end{subfigure}
	\begin{subfigure}{0.4\linewidth}
		\includegraphics[width=1\linewidth,trim=30 18 25 20, clip]{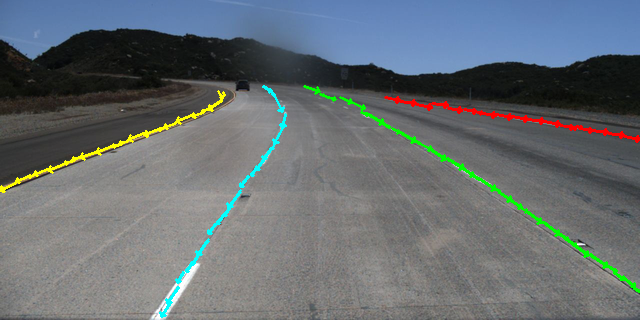}
		\caption{Breadth-first search and weighted averaging}
	\end{subfigure}
	\begin{subfigure}{0.4\linewidth}
		\includegraphics[width=1\linewidth,trim=30 18 25 20, clip]{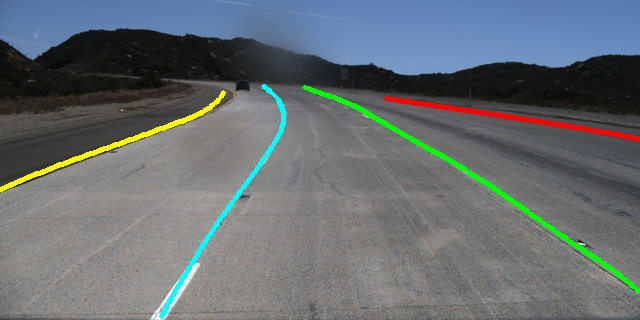}
		\caption{Spline fit}
	\end{subfigure}
	\end{center}
	\caption{Results of each step in the post-processing for TuSimple predictions. In (a), (d) and (e) the colors indicate instances. In (b) and (c) the colors describe the orientation of the line segments.}
	\label{fig:post_proc_tus}
\end{figure*}

Intermediate results of each step are depicted in \autoref{fig:post_proc_tus}.

\section{Line segment evaluation metrics}
\label{sec:scores}
For all experiments, we calculate the F1 score, recall and precision, in order to provide a general insight into the prediction independent of the application. 
We 
formulate the three scores as follows. With the number of true positive predictions $|TP|$ and the number of ground truth elements $|GT|$, the recall is defined as
\begin{equation}
    R = \frac{|TP|}{|GT|}.
\end{equation} 

Comparing the true positives with the number of predictions $|PD|$, we get a precision with 
\begin{equation}
    P = \frac{|TP|}{|PD|}.
\end{equation} 

Combining both scores leads to 
\begin{equation}
    F1 = 2 * \frac{P \cdot R}{P + R}.
\end{equation}

\begin{figure*}[tb]
	\centering
	\begin{subfigure}[t]{0.3\linewidth}
		\includegraphics[width=1\linewidth]{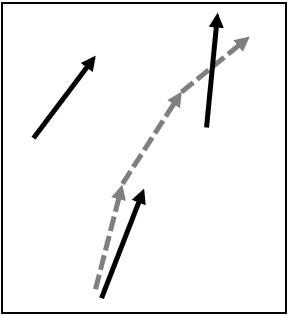}
		\caption{Exemplary predicted line segments (solid) and a single ground truth polyline (dashed).}
	\end{subfigure}
	\hfill
	\begin{subfigure}[t]{0.3\linewidth}
		\includegraphics[width=1\linewidth]{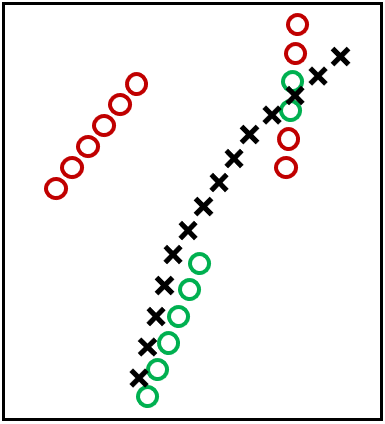}
		\caption{We calculate a precision of $P = \frac{8}{18}$. Here, red circles indicate false positives ($FP$) and green circles are accounted as true positives ($TP$).}
	\end{subfigure}
	\hfill
	\begin{subfigure}[t]{0.3\linewidth}
		\includegraphics[width=1\linewidth]{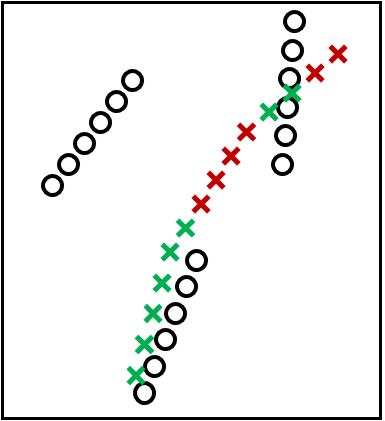}
		\caption{The recall sums up to $R = \frac{8}{14}$. We depicted false negatives ($FN$) as red crosses and true positives ($TP$) as green crosses.}
	\end{subfigure}
	\caption{Illustration of our line segment evaluation metric. Note that in the matching, not only the position of each sample, but also the angle at that position is taken into account.}
	\label{fig:eval_concept}
\end{figure*}

\begin{figure*}[tb]
    \begin{center}	
		\begin{subfigure}{0.49\linewidth}
			\centering
			\includegraphics[width=1\linewidth,trim=35 200 600 200, clip]{res/\tus/tusimple/0530_1492627321448785174_0.png} 
			\caption{Ground truth}
		\end{subfigure}
		\begin{subfigure}{0.49\linewidth}
		    \centering
		    \includegraphics[width=1\linewidth,trim=30 18 25 20, clip]{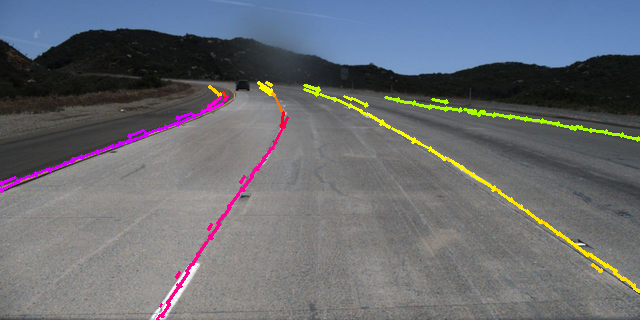}
			\caption{Prediction}
		\end{subfigure}
		\begin{subfigure}{0.49\linewidth}
		    \centering
		    \includegraphics[width=1\linewidth]{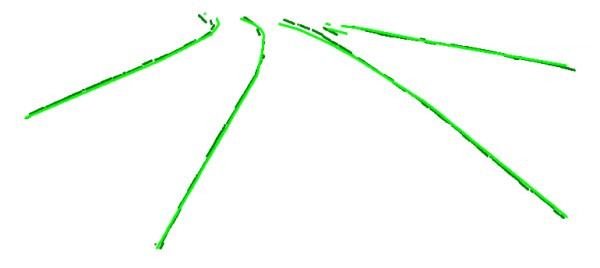}
			\caption{Only all true positive matches with prediction in dark and ground truth in light green.}
		\end{subfigure}
		\begin{subfigure}{0.49\linewidth}
		    \centering
		    \includegraphics[width=1\linewidth]{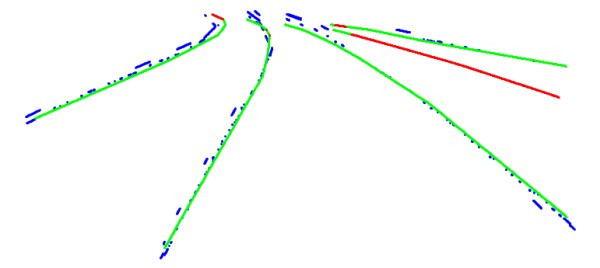}
			\caption{Unmatched predictions (false positives, blue) and unmatched ground truth (false negatives, red).}
		\end{subfigure}
	\end{center}
	\caption{Visualization of the true positives, false positives and false negatives for a real prediction. Here, recall $R=0.81$ and precision $P=0.86$.}
	\label{fig:tus_fp_fn}
\end{figure*}

As in our representation, line segments can differ in length and orientation, determining the true positives should regard these parameters. We thus sample all line segments both from the predictions and from the ground truth with a sample distance of \SI{1}{px}. In addition, we calculate the orientation $\alpha$ of each line segment, leading to a representation of each sample point as $s=(x,y,\alpha)$. Then, we account all predictions as true positive that lie within a certain radius $\theta$ from a given ground truth point. \autoref{fig:eval_concept} visualizes this for the 2D case without regarding the orientation $\alpha$. An example from the TuSimple dataset is depicted in \autoref{fig:tus_fp_fn}.


\section{Relevant experiments}
\label{sec:results}
For better comparison between the several ablation studies, we provide all results in \autoref{tab:all_experiments}. The first group depicts experiments for different line representations, the second group provides insight into the grid resolution and in the third group, you can compare different numbers of predictors. 

\begin{table*}[tb]
	\centering
	\begin{tabular}{|rrr||lll|llll|}
    \hline
      Line        & Pred  & Grid    & Acc 			& FP  	       & FN 		  & F1 		     & Recall        & Prec.          &      \\ \hline \hline
    \acs{lrep1d}  & 8     & 32\,px  & .887 	& \snd{.157}   & .16	  & .616         & \snd{.955}    & .455           &         \\ %
    \acs{lrepeu}  & 8     & 32\,px  & .877          & \first{.137} & .168         & .685   & \snd{.956}  & .533     &         \\ %
    \acs{lreppo}  & 8     & 32\,px  & \snd{.914}  & \snd{.159}         & \snd{.112} & \snd{.740} & \snd{.950}          & \snd{.607}   &         \\ %
    \hline

    \hline
    \acs{lrepeu}  & 8     & 32\,px  & .877          & \first{.137}   & .168         & .685         & \snd{.956}    & .533           &         \\ %
    \acs{lrepeu}  & 8     & 16\,px  & .899          & .343         & .201   	  & .658       	 & \snd{.959}  & .500           &      \\ %
    \acs{lrepeu}  & 8     & 8\,px   & \snd{.839}          & .481         & .331  	      & .712         & .919          & .581           &      \\ %
    \hline
    \acs{lreppo}  & 8     & 32\,px  & \snd{.914}          & \snd{.159}         & \snd{.112}   & {.74}        & .948          & \snd{.607}           &         \\ %
    \acs{lreppo}  & 8     & 16\,px  & \first{.930}  & .258         & \first{.095} & \snd{.739}         & \snd{.951}          & \snd{.604}           &          \\ %
    \acs{lreppo}  & 8     & 8\,px   & .875          & .405         & .196         & \first{.776}         & .930          & \first{.665}           &         \\ %
	\hline

    \hline
    \acs{lrepeu}    & 4     & 32\,px  & .878          & .168         & .180          & .546         & \first{.971}  & .380           &           \\ %
    \acs{lrepeu}    & 8     & 32\,px  & .877          & \first{.137}   & .168         & .685         & \snd{.956}    & .533   	   &           \\ %
    \acs{lrepeu}    & 12    & 32\,px  & .885          & \snd{.154}         & .159         & .717         & .940           & .580           &           \\ \hline %
    \acs{lreppo}    & 4     & 32\,px  & \first{.922}  & \snd{.157}         & \first{.099} & .713         & \snd{.952}          & .571 		   &           \\ %
    \acs{lreppo}    & 8     & 32\,px  & \snd{.914}    & \snd{.159}         & \snd{.112}   & \snd{.74}    & .948          & \snd{.607}    &           \\ %
    \acs{lreppo}    & 12    & 32\,px  & .903          & \first{.135} & .120         & \first{.777} & .941          & \first{.662}  &           \\ %
    \hline

  \end{tabular}
    \caption{Results for the ablation studies on important hyperparameters on the Tusimple validation set. Best result in each group is bold, second best result is underlined.}
    \label{tab:all_experiments}
\end{table*} %

\onecolumn
\clearpage
\clearpage

\section{Qualitative results for TuSimple}
\label{sec:tus}

\begin{figure*}[htb!]
	\begin{center}	
		\begin{subfigure}{0.49\linewidth}
			\centering
			\includegraphics[width=1\linewidth,trim=35 200 600 200, clip]{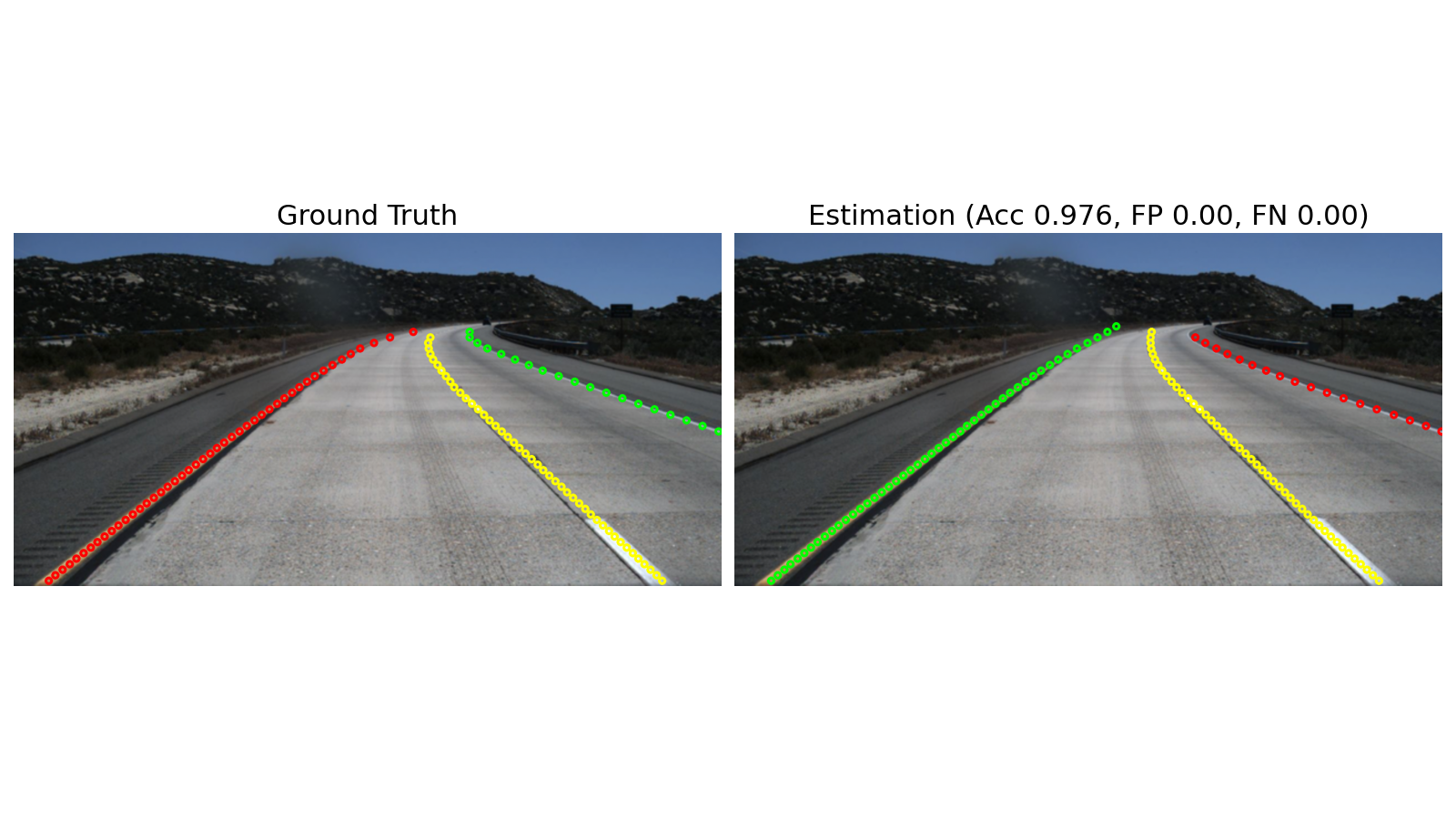}
			\includegraphics[width=1\linewidth,trim=35 200 600 200, clip]{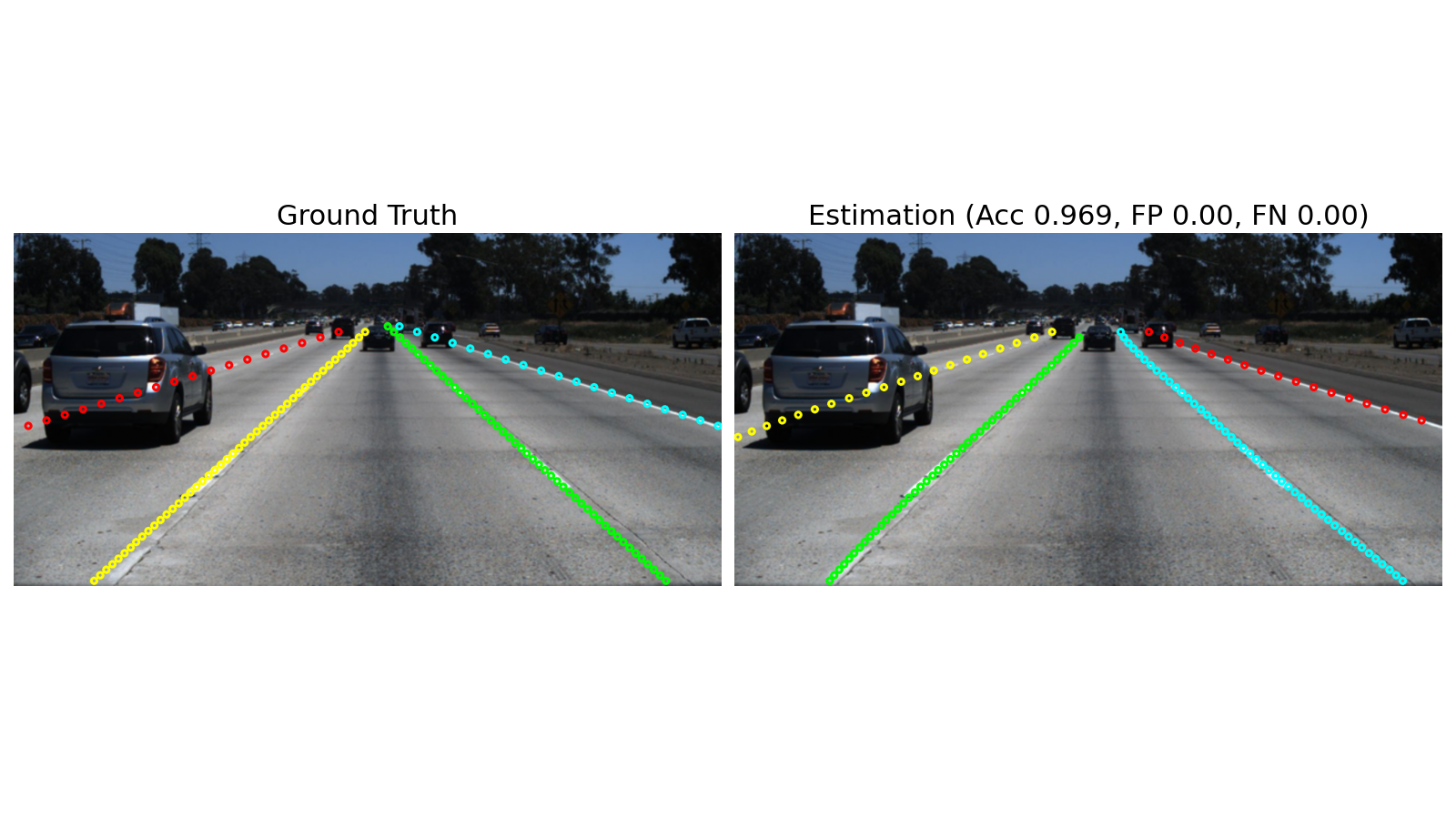}
			\includegraphics[width=1\linewidth,trim=35 200 600 200, clip]{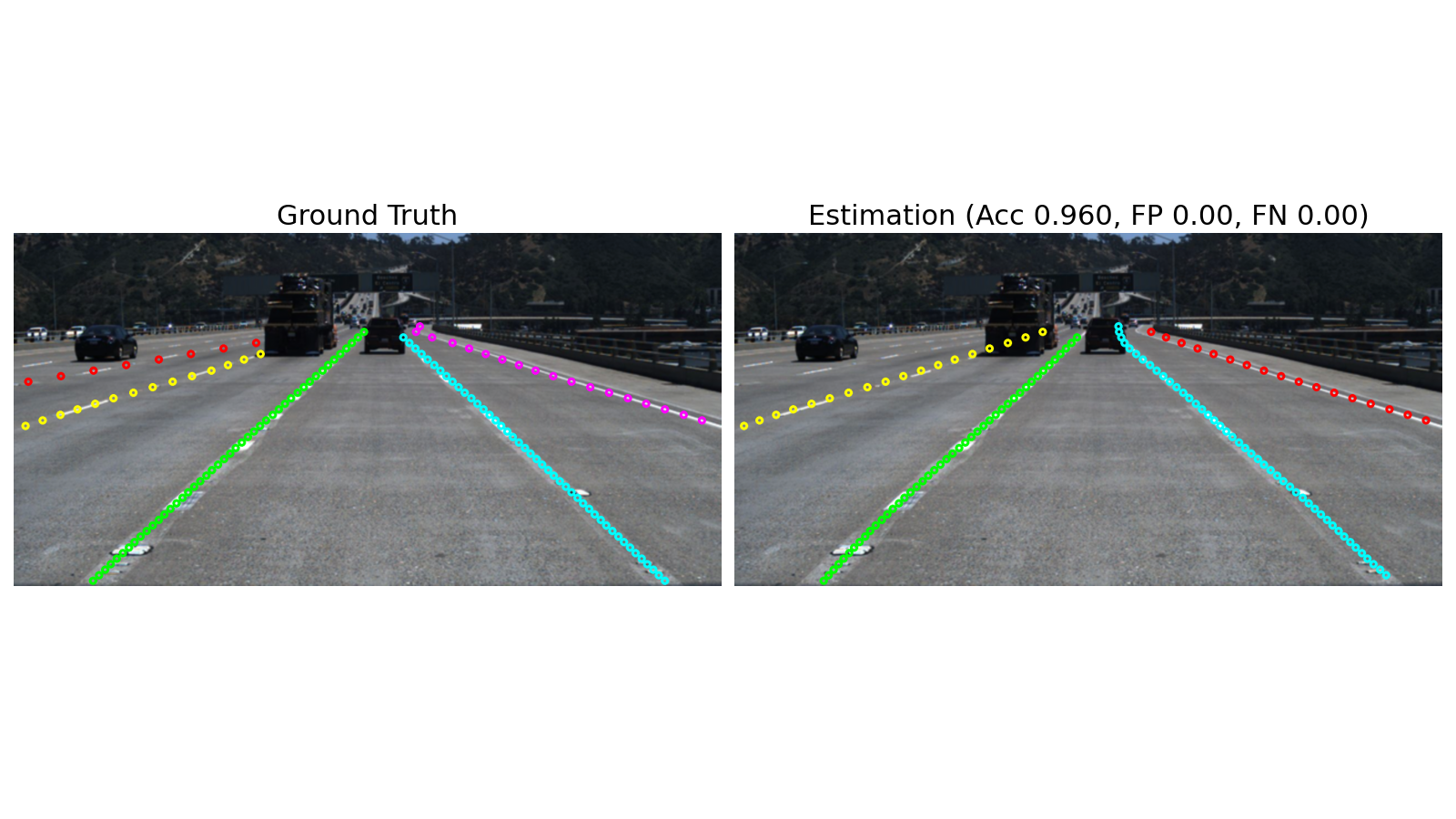}
			\includegraphics[width=1\linewidth,trim=35 200 600 200, clip]{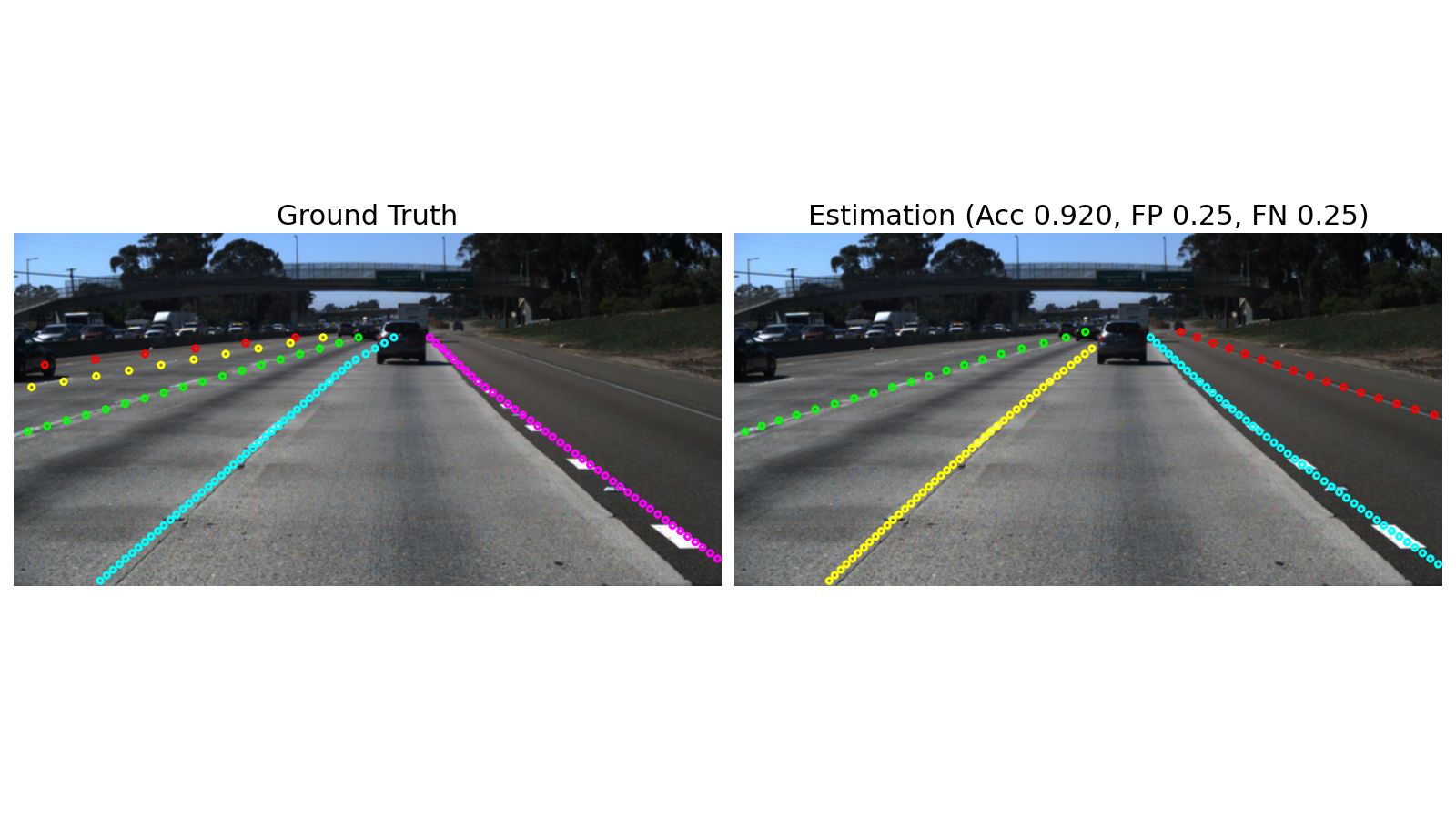}
			\includegraphics[width=1\linewidth,trim=35 200 600 200, clip]{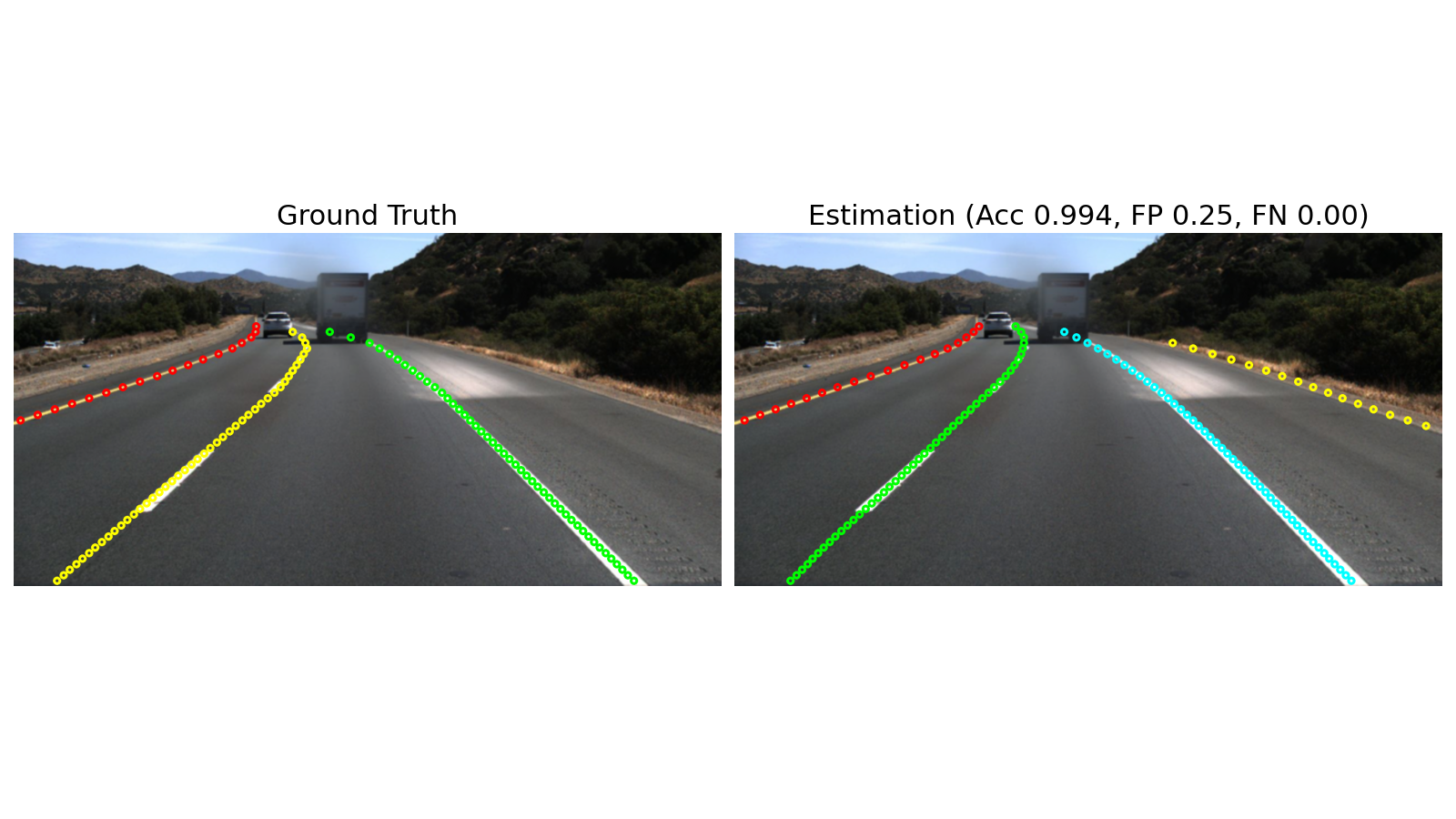}
			\caption{Ground Truth}
		\end{subfigure}
		\begin{subfigure}{0.49\linewidth}
		    \centering

		    \includegraphics[width=1\linewidth,trim=30 18 25 20, clip]{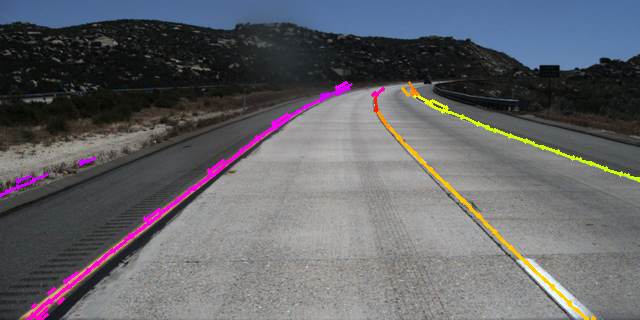}
            \includegraphics[width=1\linewidth,trim=30 18 25 20, clip]{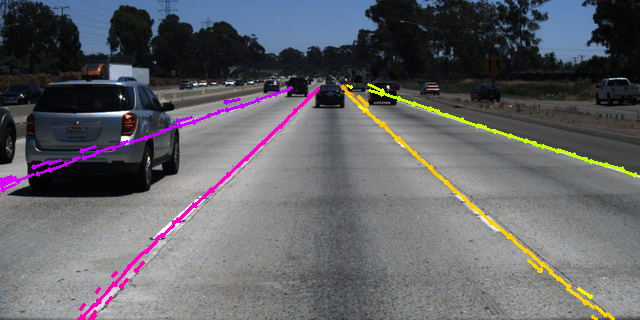}
            \includegraphics[width=1\linewidth,trim=30 18 25 20, clip]{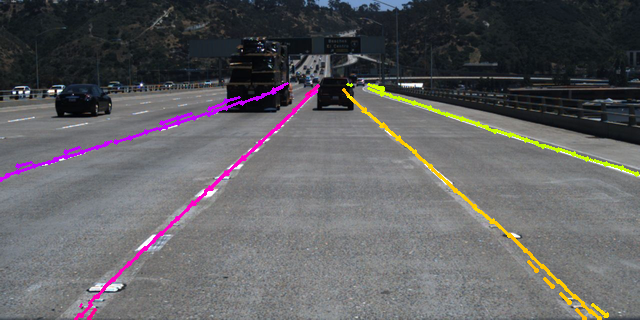}
            \includegraphics[width=1\linewidth,trim=30 18 25 20, clip]{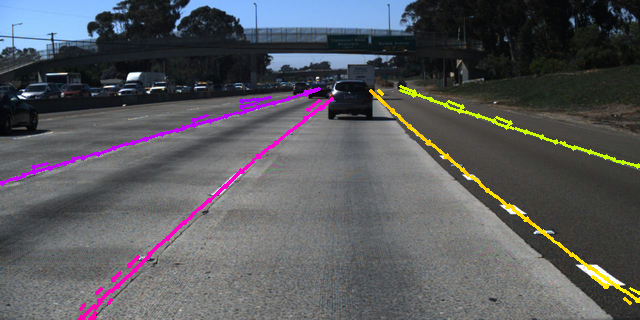}
		    \includegraphics[width=1\linewidth,trim=30 18 25 20, clip]{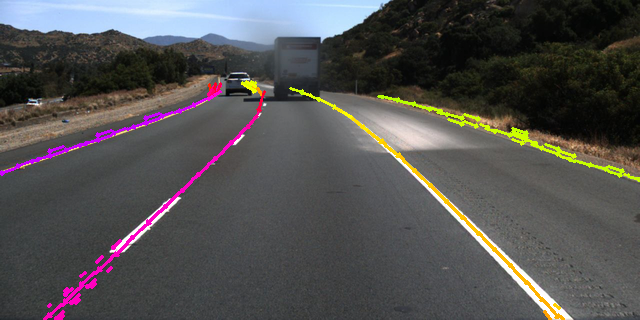}
			\caption{Prediction after generic NMS}
		\end{subfigure}
	\end{center}
	\caption{Example results of the TuSimple test set (\acf{lreppo}, eight predictors and a grid resolution of 16$\times$16\,px). Colors in (a) indicate instances. In (b) the colors visualize the orientation of the predicted line segments.}
	\label{fig:tus_results2}
\end{figure*}

We depicted further examples for the TuSimple dataset in \autoref{fig:tus_results} and \autoref{fig:tus_results2}. These examples clearly show some peculiarities of the TuSimple dataset and training. First, the prediction often shows two highly confident hypotheses on the lane boundary, which is probably caused by imprecise labels and ambiguities within. As can be seen in the visualization of the main submission, we are easily discarding these false hypotheses in the post-processing.

Similarly, predicting false positive and false negative lane boundaries happens (FP rate of .188 and FN rate of .076) as there also exist quite some ambiguities in the labeling. E.g. considering the
first row
in \autoref{fig:tus_results}, the blue lane boundary (in the ground truth) does not have any visual cue and is very doubtful even for humans. Thus, our approach does not recognize this as a valid boundary. 
Comparing the 
second and 
third row 
of \autoref{fig:tus_results}, it becomes apparent why a learned architecture might show frequent false positives or false negatives. In the 
second 
row, five lane boundaries were labelled whereas the 
third 
row only expects three lane boundaries although there exists at least four (+1 next to the right green lane boundary). The same applies to the 
fourth row 
of \autoref{fig:tus_results2}, where three lanes to the left are labelled, but the one to the right is not. 
This leads to false positive predictions by our approach that predicts valid lane boundary hypotheses that are not labelled within TuSimple dataset. 

Further, we found scenes, where the ground truth is wrong, but our prediction overcomes those errors e.g. as shown in the 
third row 
of \autoref{fig:tus_results2}.

\begin{figure*}[htb!]
	\begin{center}	
		\begin{subfigure}{0.49\linewidth}
			\centering
			\includegraphics[width=1\linewidth,trim=35 200 600 200, clip]{res/\tus/tusimple/0530_1492627321448785174_0.png}
			\centering
			\includegraphics[width=1\linewidth,trim=35 200 600 200, clip]{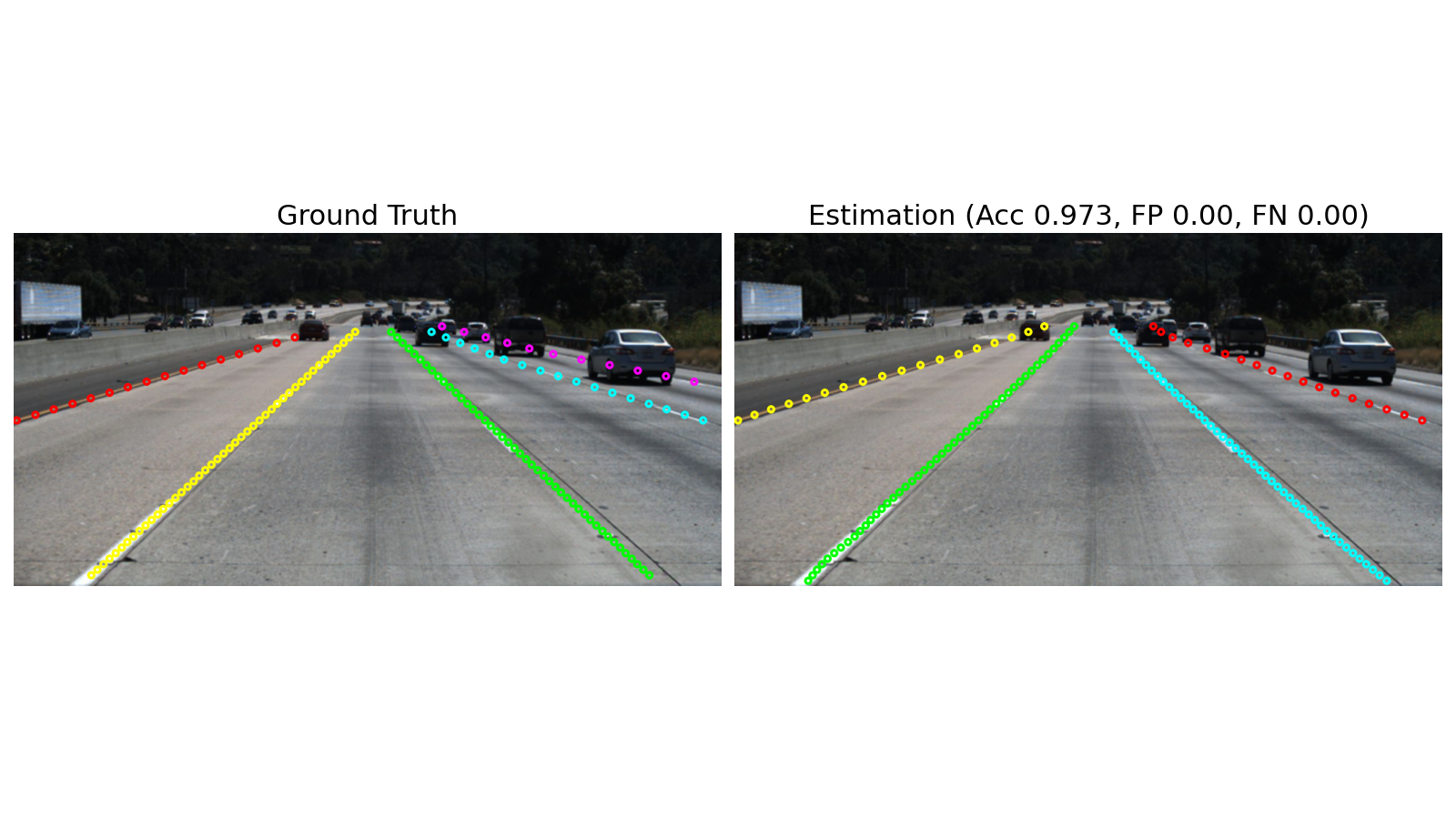}
			\centering
			\includegraphics[width=1\linewidth,trim=35 200 600 200, clip]{res/\tus/tusimple/0530_1492638943713602995_0.png} %
			\centering
			\caption{Ground Truth}
		\end{subfigure}
		\begin{subfigure}{0.49\linewidth}
		    \centering
            \includegraphics[width=1\linewidth,trim=30 18 25 20, clip]{res/\tus/points/0530_1492627321448785174_0_3_nms.png}
            \centering
            \includegraphics[width=1\linewidth,trim=30 18 25 20, clip]{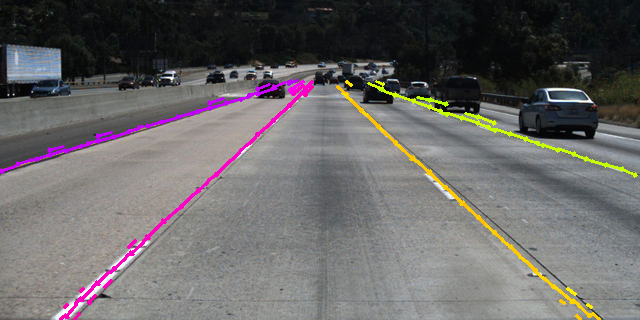}
            \centering
            \includegraphics[width=1\linewidth,trim=30 18 25 20, clip]{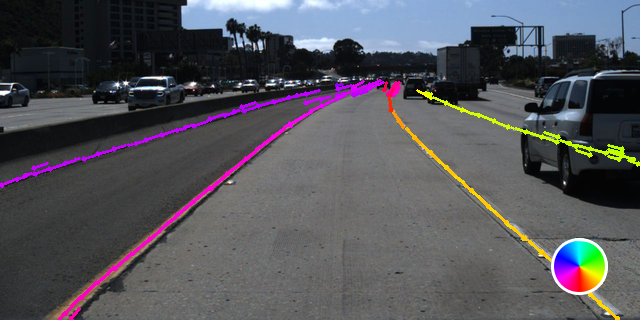}
			\caption{Prediction after generic NMS}
		\end{subfigure}
	\end{center}
	\caption{Example results of the TuSimple test set (\acf{lreppo}, eight predictors and a grid resolution of 16$\times$16\,px). Colors in (a) indicate instances. In (b) the colors visualize the orientation of the predicted line segments.}
	\label{fig:tus_results}
\end{figure*}
\clearpage

\clearpage %
\section{Qualitative results for \acl{KAI}}
\label{sec:aer}

\begin{figure*}[htb!]
	\begin{center}	
		\begin{subfigure}{0.309\linewidth}
			\centering 			
 			\includegraphics[width=1\linewidth]{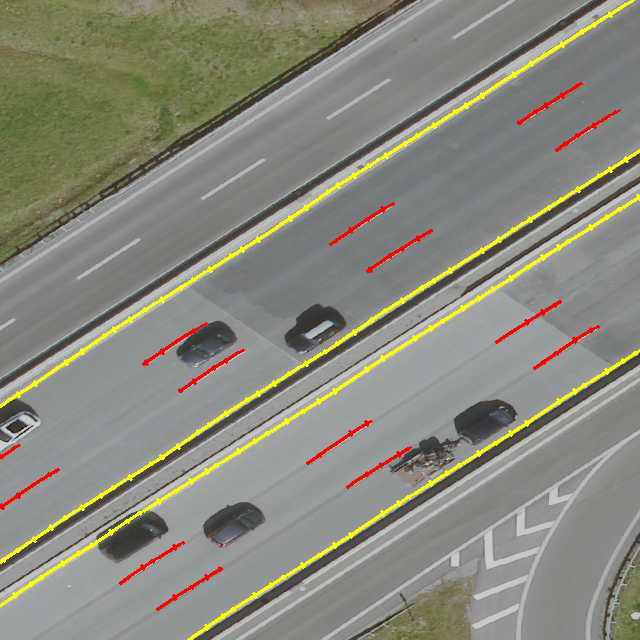}  
 			\includegraphics[width=1\linewidth]{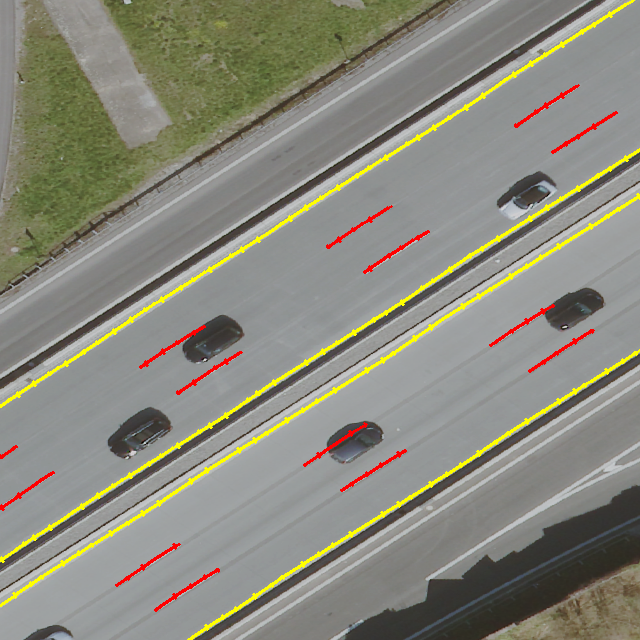}  
			\caption{Ground truth}
		\end{subfigure}
		\begin{subfigure}{0.309\linewidth}
		    \centering 			
 			\includegraphics[width=1\linewidth]{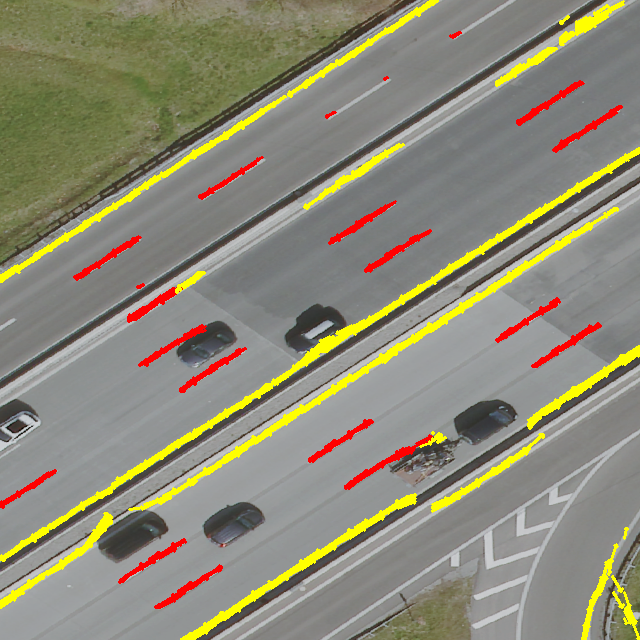}
 			\includegraphics[width=1\linewidth]{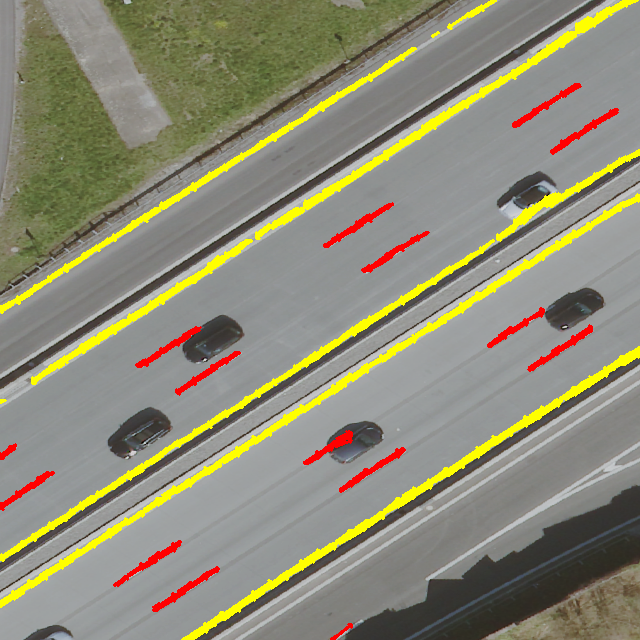} 
			\caption{Prediction}
		\end{subfigure}
		\begin{subfigure}{0.309\linewidth}
		    \centering
 			\includegraphics[width=1\linewidth]{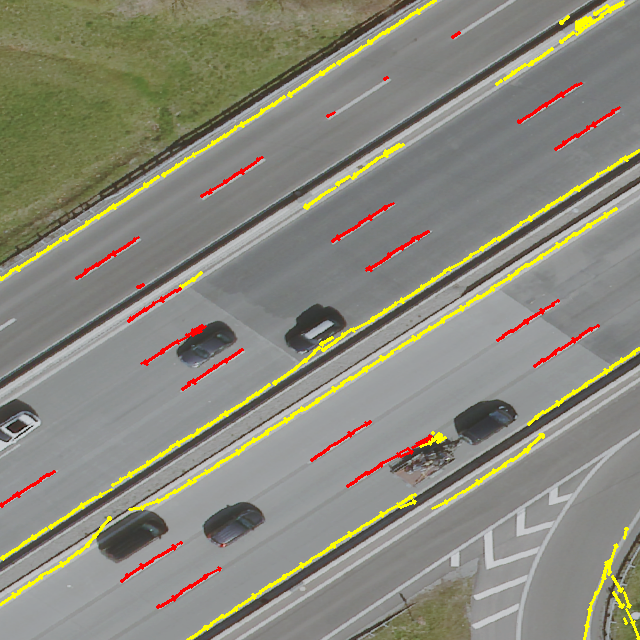} 
 			\includegraphics[width=1\linewidth]{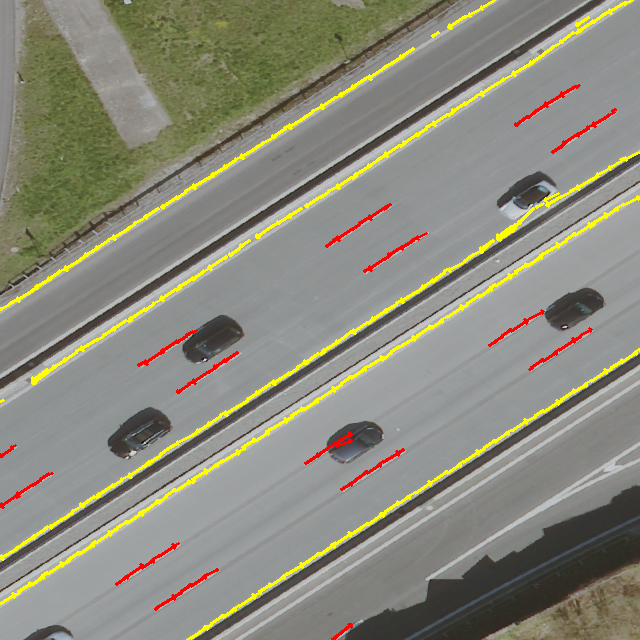} 
			\caption{Prediction after NMS}
		\end{subfigure}
	\end{center}
	\caption{Example results of the KAI dataset (\acf{lreppo}, eight predictors and a grid resolution of 16$\times$16\,px).\\
	Aerial images: \textcopyright~City of Karlsruhe $\vert$ Liegenschaftsamt}
	\label{fig:aer1_results}
\end{figure*}

\begin{figure*}[htb!]
	\begin{center}	
		\begin{subfigure}{0.3\linewidth}
			\centering 			
 			\includegraphics[width=1\linewidth]{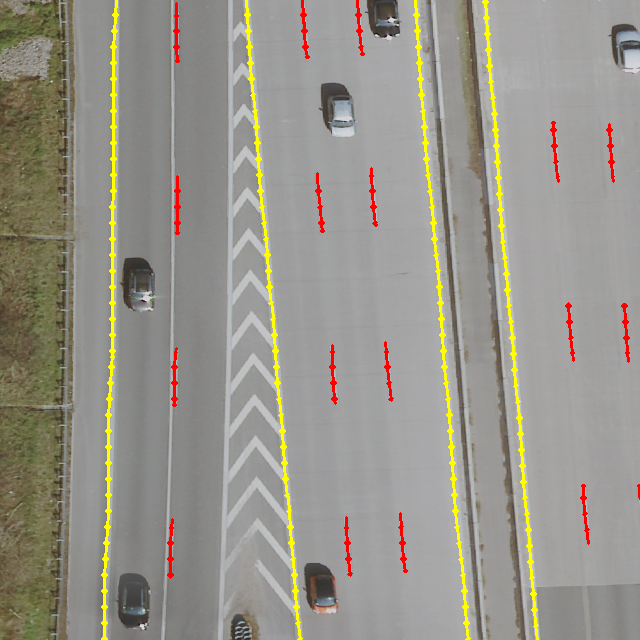} 
 			\includegraphics[width=1\linewidth]{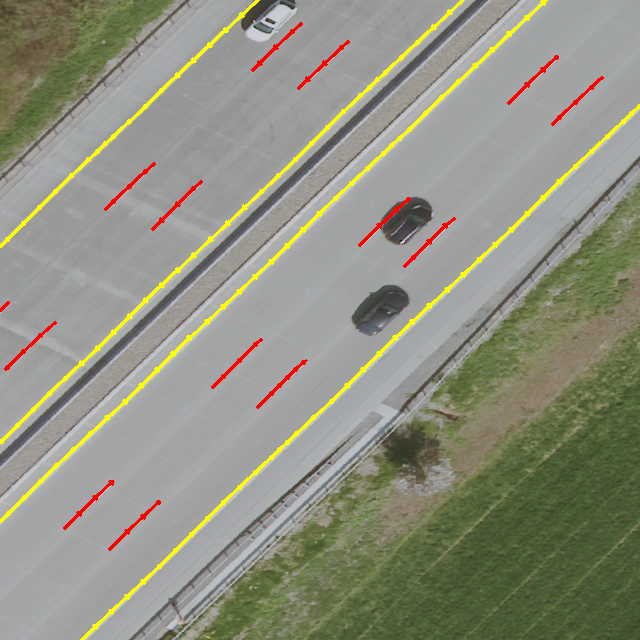} 
 			\includegraphics[width=1\linewidth]{{res/\aerial/points/image_1341_sn_48.9990934_8.4446478_1_label}.png} 
 			\includegraphics[width=1\linewidth]{{res/\aerial/points/image_11_ns_49.0469014_8.4918172_1_label}.png} 
			\caption{Ground Truth}
		\end{subfigure}
		\begin{subfigure}{0.3\linewidth}
		    \centering 			  
 			\includegraphics[width=1\linewidth]{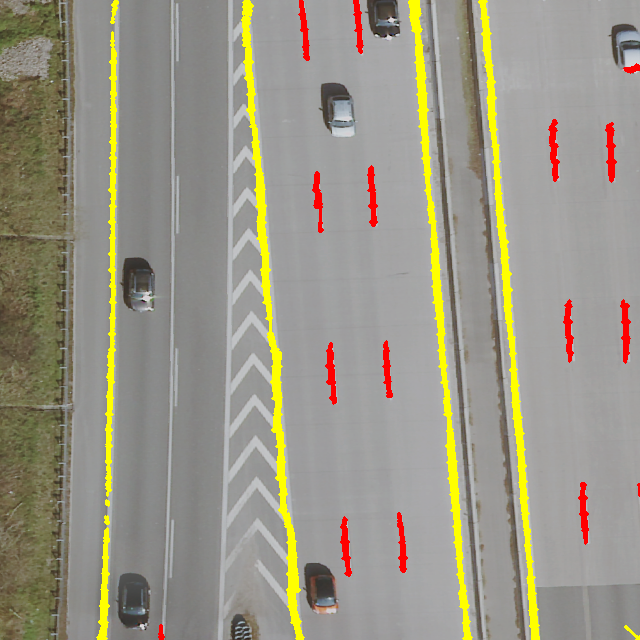} 
 			\includegraphics[width=1\linewidth]{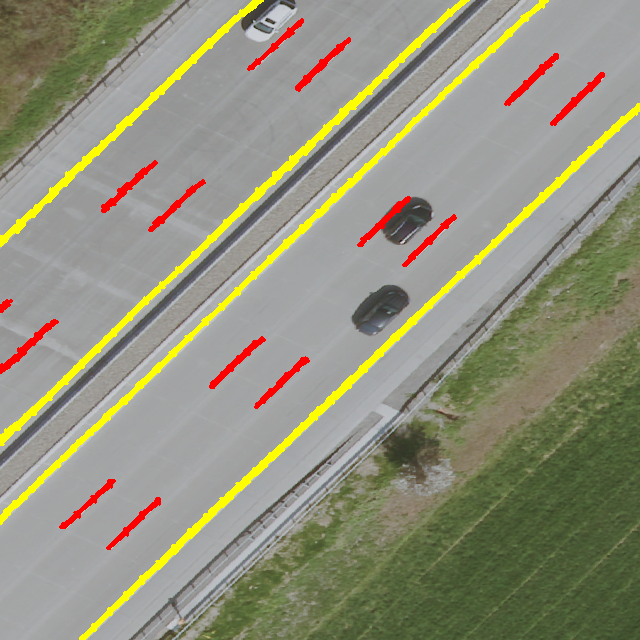} 
 			\includegraphics[width=1\linewidth]{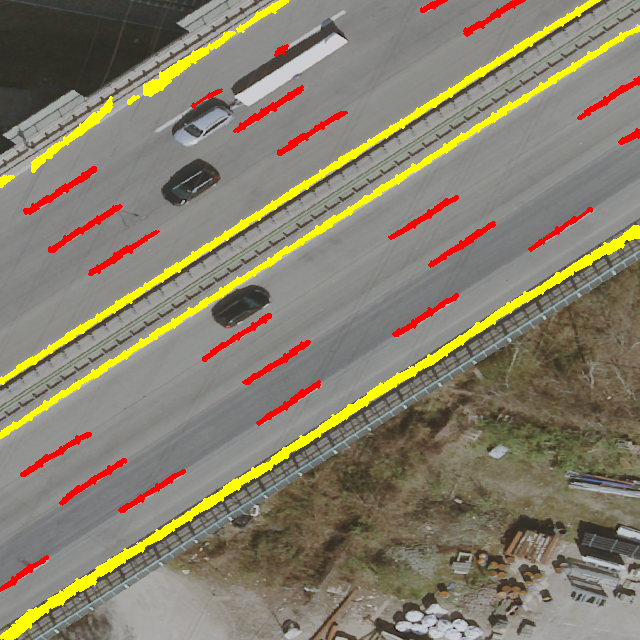} 
 			\includegraphics[width=1\linewidth]{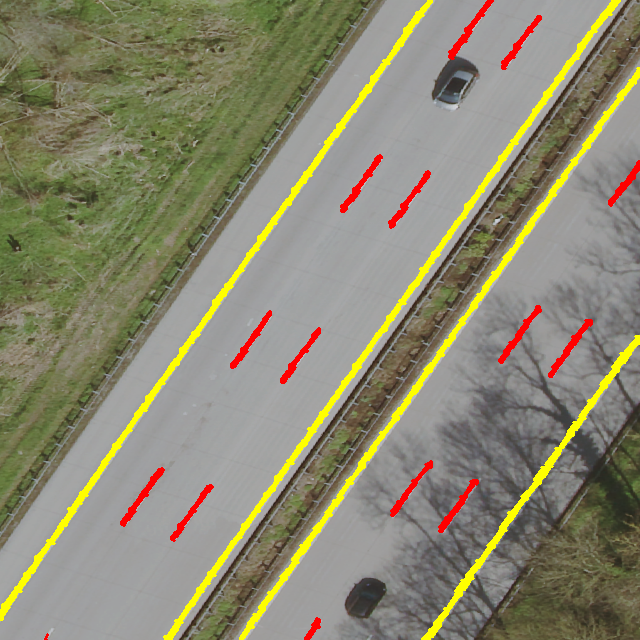} 
			\caption{Prediction}
		\end{subfigure}
		\begin{subfigure}{0.3\linewidth}
		    \centering
 			\includegraphics[width=1\linewidth]{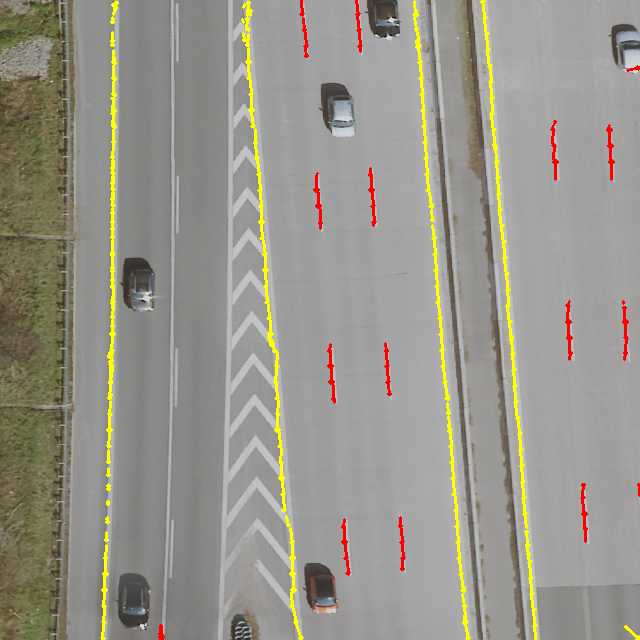} 
 			\includegraphics[width=1\linewidth]{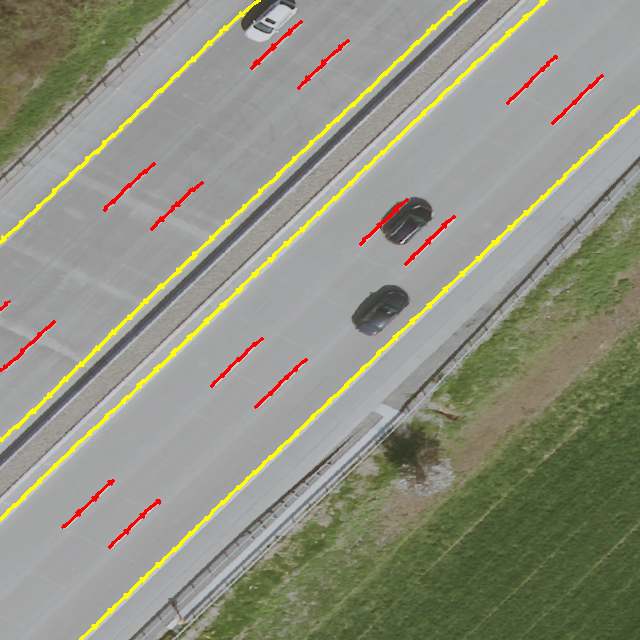} 
 			\includegraphics[width=1\linewidth]{{res/\aerial/points/image_1341_sn_48.9990934_8.4446478_3_nms}.png} 
 			\includegraphics[width=1\linewidth]{{res/\aerial/points/image_11_ns_49.0469014_8.4918172_3_nms}.png} 
			\caption{Prediction after NMS}
		\end{subfigure}
	\end{center}
	\caption{Example results of the KAI dataset (\acf{lreppo}, eight predictors and a grid resolution of 16$\times$16\,px).\\Aerial images: \textcopyright~City of Karlsruhe $\vert$ Liegenschaftsamt}
	\label{fig:aer2_results}
\end{figure*}

We depicted further examples for the \acs{KAI} dataset in \autoref{fig:aer1_results} and \autoref{fig:aer2_results}. As can be seen, we are able to classify the road markings with \SI{96}{\percent} accuracy. However, the geometry of the prediction provides more valuable insights. 

As can be seen in \autoref{fig:aer1_results}, we are able to predict unusual side lanes that are not in the ground truth. However, some of the dashed markings are predicted only partly. In the first row of \autoref{fig:aer2_results}, we can see that the dataset bears some unexpected pattern in the labels. For non-drivable areas only the right lane marking is labeled. This is learned by our network, but is not the best in general. In the same image, we can see that unusual road marking combinations (dashed lane marking adjacent to solid lane marking) are not recognized. This can be explained by the small number of occurrences in the training set. 

In the first row of \autoref{fig:aer1_results} and in the first and third row of \autoref{fig:aer2_results}, we can further see that unusual and especially white vehicles confuse the estimation slightly. This might also be due to their small representation in the dataset. 

On the contrary, shadows and artifacts on the surface seem to have no effect on the prediction (cf. second and fourth rows of \autoref{fig:aer2_results})

\clearpage %

\section{Qualitative results for Argoverse}
\label{sec:argo}

The Argoverse datasets provides the most interesting insights into the capabilities of YOLinO. We primarily selected intersection scenes in order to show that YOLinO is able to infer the complex structure. The representation is ideally designed to describe multiple directions and overlapping lanes. 

\begin{sidewaysfigure*}[htb!]
	\begin{center}	
		\begin{subfigure}{0.309\linewidth}
			\centering 			
 			\includegraphics[width=1\linewidth]{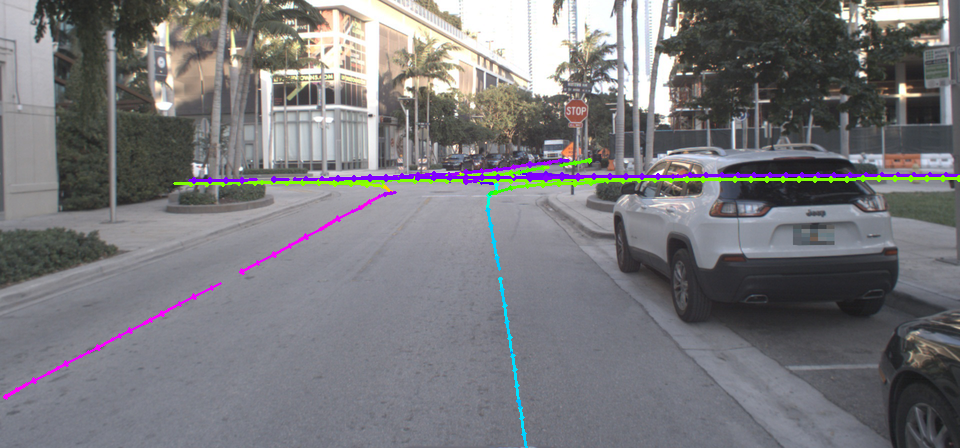} 
 			\includegraphics[width=1\linewidth]{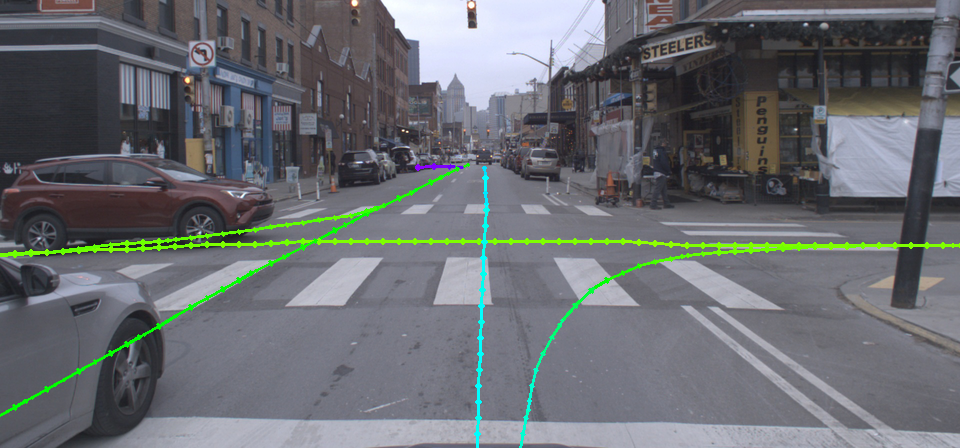} 
 			\includegraphics[width=1\linewidth]{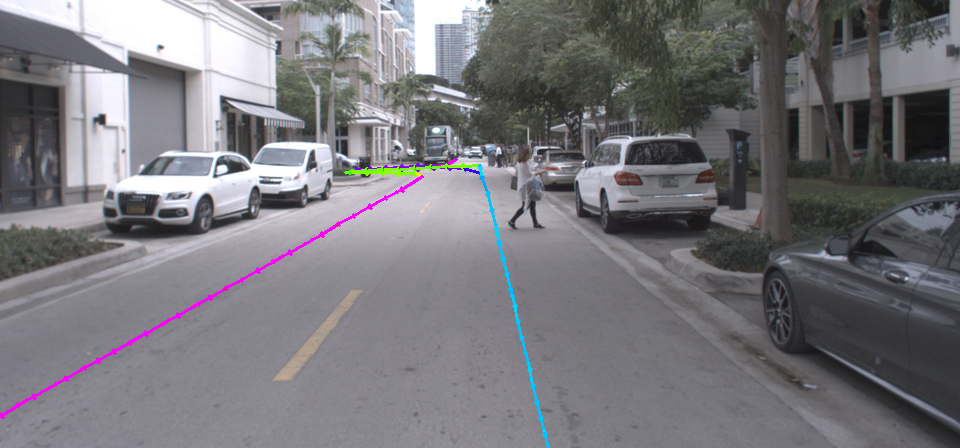} 
 			\includegraphics[width=1\linewidth]{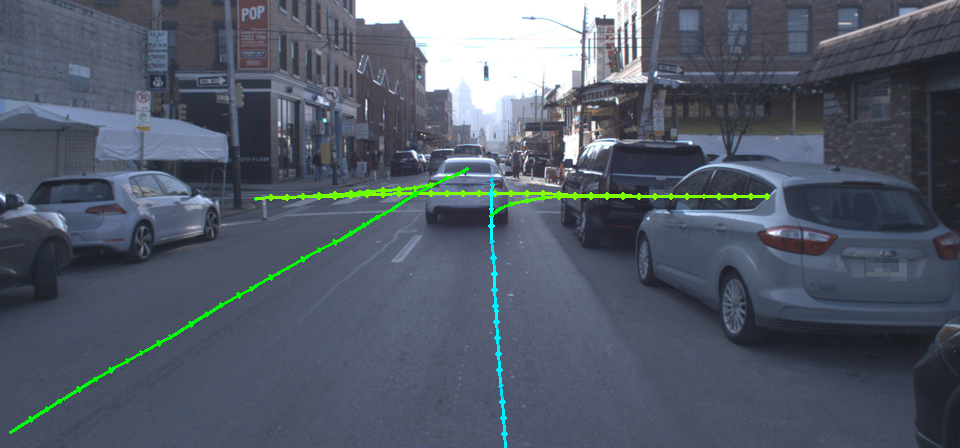}  
			\caption{Ground Truth}
		\end{subfigure}
		\begin{subfigure}{0.309\linewidth}
		    \centering 			
 			\includegraphics[width=1\linewidth]{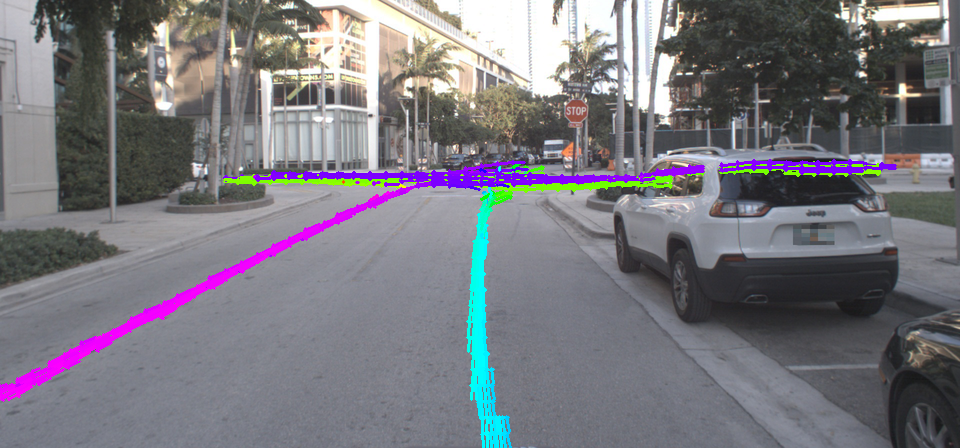} 
 			\includegraphics[width=1\linewidth]{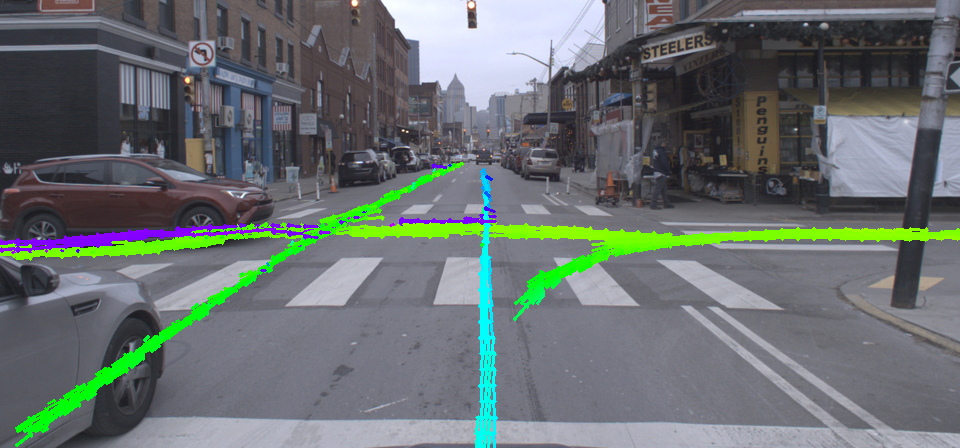} 
 			\includegraphics[width=1\linewidth]{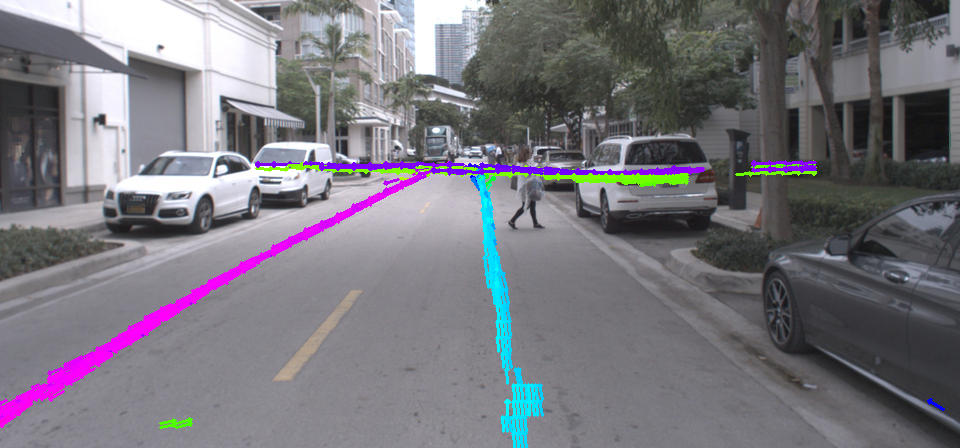} 
 			\includegraphics[width=1\linewidth]{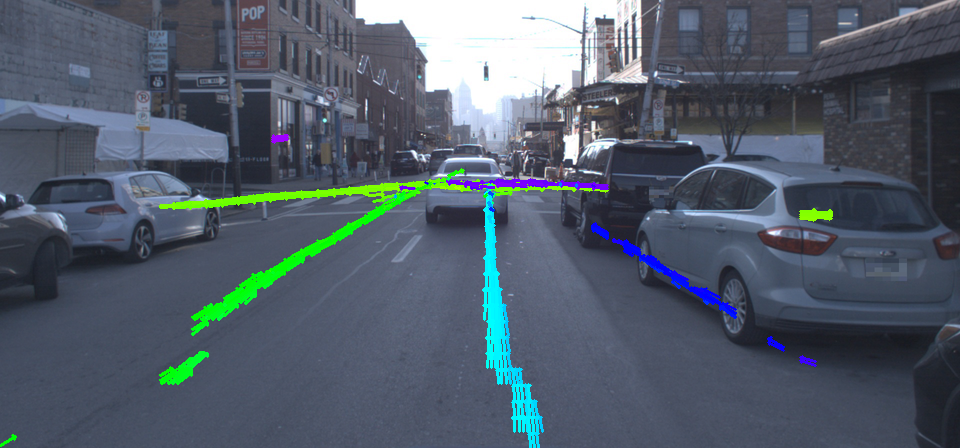}  
			\caption{Prediction}
		\end{subfigure}
		\begin{subfigure}{0.309\linewidth}
		    \centering
 			\includegraphics[width=1\linewidth]{res/\argo/points/d5_l18_i280_3_nms.png} 
 			\includegraphics[width=1\linewidth]{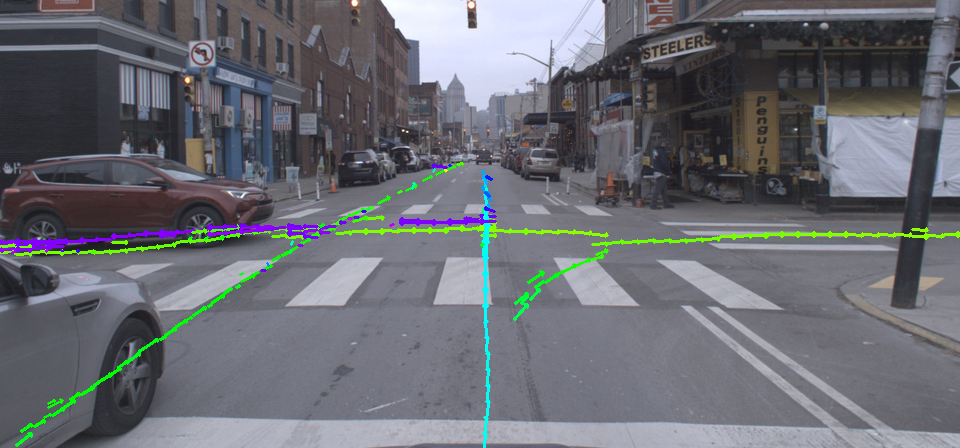} 
 			\includegraphics[width=1\linewidth]{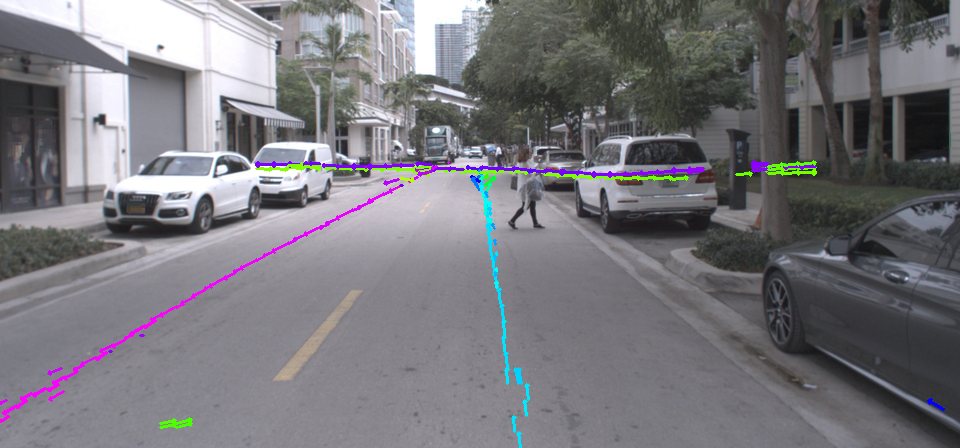} 
 			\includegraphics[width=1\linewidth]{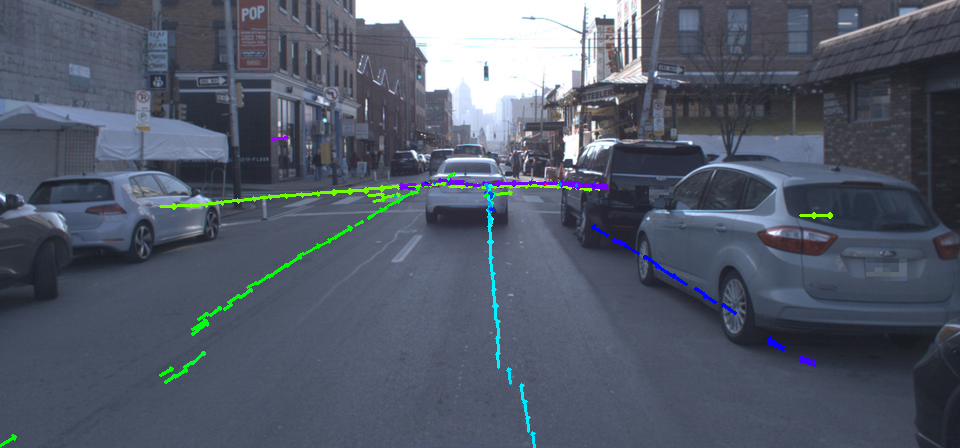}  
			\caption{Prediction post-processed}
		\end{subfigure}
	\end{center}
	\caption{Example results of the Argoverse dataset (\acf{lreppo}, eight predictors and a grid resolution of 16$\times$16\,px). Colors indicate the orientation of the predicted line segments.}
	\label{fig:argo2_results}
\end{sidewaysfigure*}

\Cref{fig:argo_results} shows several scenes where mainly straight lane centerlines are predicted. Here, no traffic participant is occupying the view, thus the prediction is straight forward. On the contrary, in \autoref{fig:argo2_results} parts of the road are hidden behind other vehicles or pedestrians are crossing the street. This seems not to be of any problem for the approach. For both, we want to highlight that centerlines on empty streets might already be a tough detection target as they are not determined by direct visual cues. 

Comparing the second row of \autoref{fig:argo_results} with the second and fourth rows of \autoref{fig:argo2_results}, it becomes clear that the network is also able to correctly predict the driving direction of lanes, as \autoref{fig:argo2_results} also shows scenes with a single driving direction.  

In some scenes (2nd row in \autoref{fig:argo2_results}, 2nd row in \autoref{fig:argo_results}), the network assumes a three-armed intersection to be a four-way intersection. This only occurs in far distance and might be a result of an imbalance in the dataset. 
Surprisingly, the scene in the last row of \autoref{fig:argo2_results} seems to pose more problems than others as the left lane is not fully detected, and more false positives than usual are predicted. This might be due to difficult lighting conditions. 
\begin{sidewaysfigure*}[htb!]
	\begin{center}	
		\begin{subfigure}{0.309\linewidth}
			\centering 			
 			\includegraphics[width=1\linewidth]{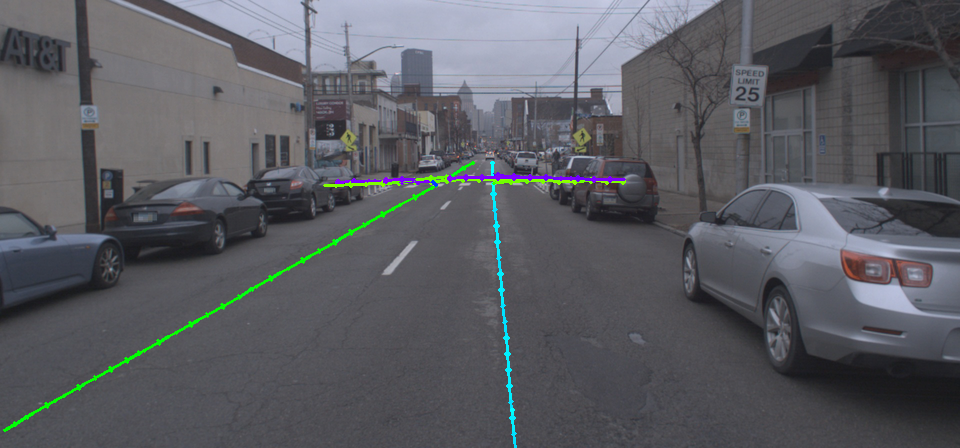} 
 			\includegraphics[width=1\linewidth]{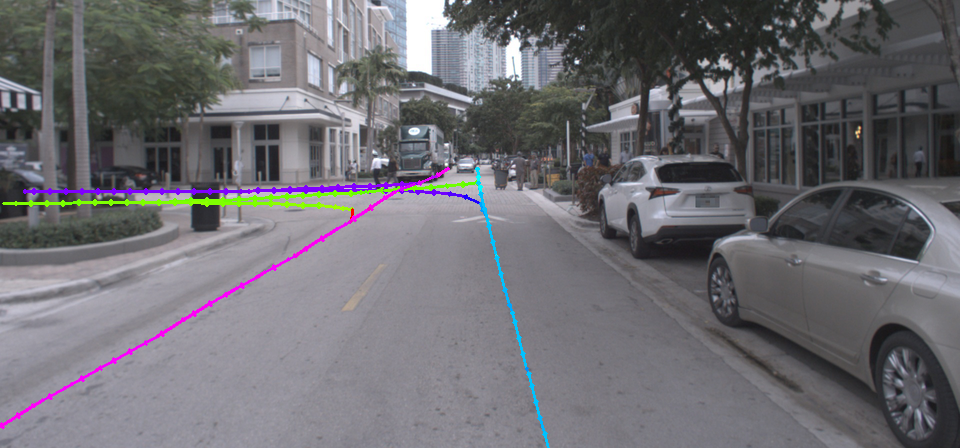} 
 			\includegraphics[width=1\linewidth]{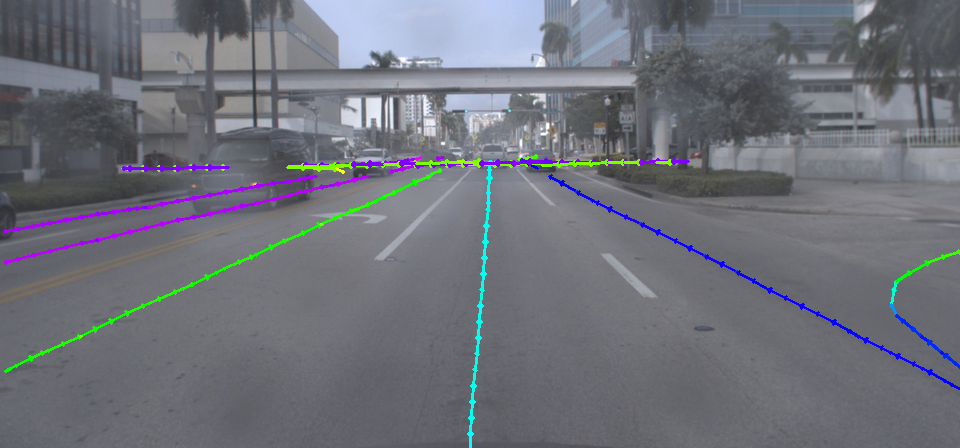} 
			\caption{Ground Truth}
		\end{subfigure}
		\begin{subfigure}{0.309\linewidth}
		    \centering 			
 			\includegraphics[width=1\linewidth]{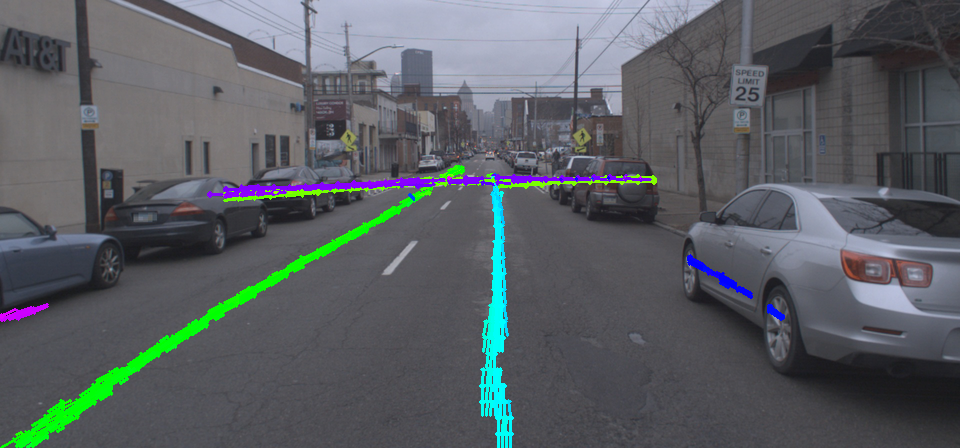} 
 			\includegraphics[width=1\linewidth]{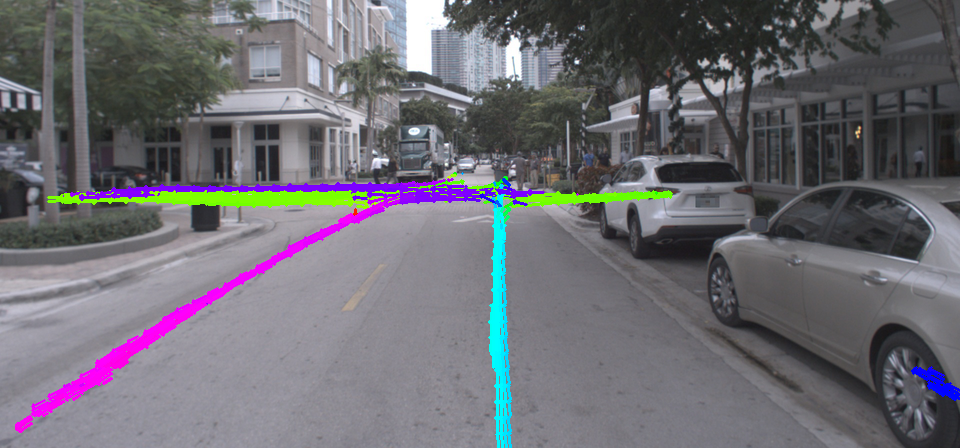} 
 			\includegraphics[width=1\linewidth]{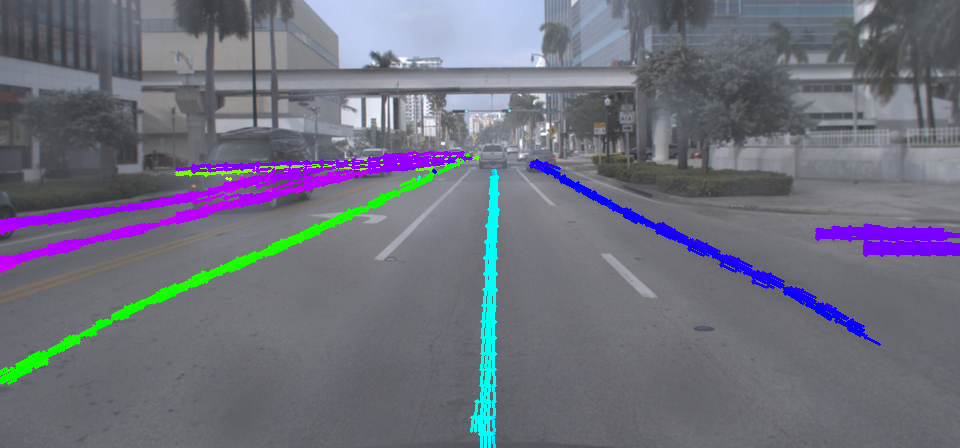} 
			\caption{Prediction}
		\end{subfigure}
		\begin{subfigure}{0.309\linewidth}
		    \centering
 			\includegraphics[width=1\linewidth]{res/\argo/points/d5_l19_i15_3_nms.png} 
 			\includegraphics[width=1\linewidth]{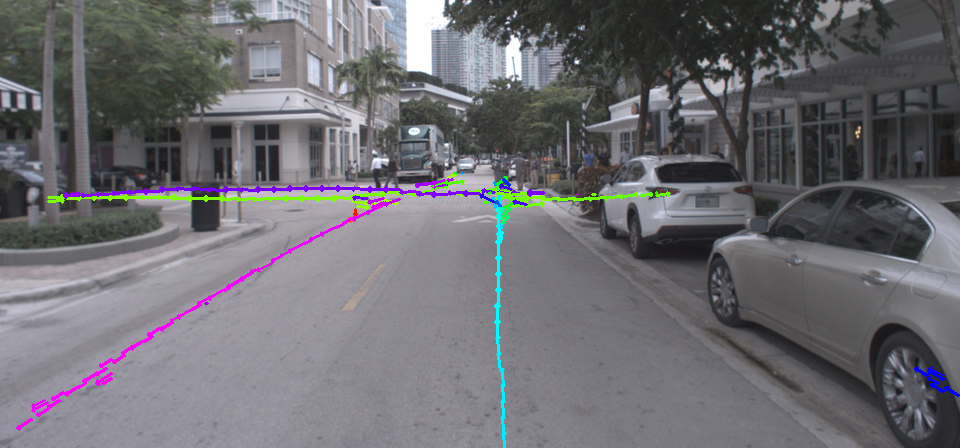} 
 			\includegraphics[width=1\linewidth]{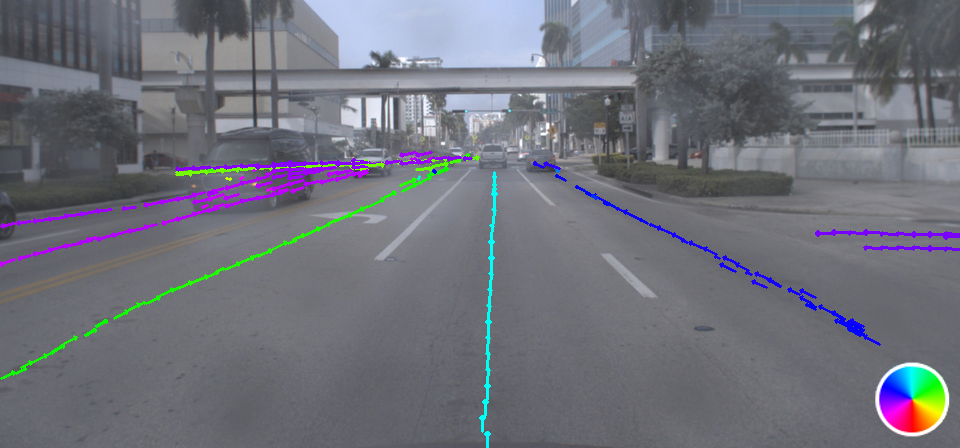} 
			\caption{Prediction post-processed}
		\end{subfigure}
	\end{center}
	\caption{Example results of the Argoverse dataset (\acf{lreppo}, eight predictors and a grid resolution of 16$\times$16\,px). Colors indicate the orientation of the predicted line segments.}
	\label{fig:argo_results}
\end{sidewaysfigure*}

\clearpage

\clearpage 
\twocolumn
{\small
\bibliographystyle{ieee_fullname}
\bibliography{bibliography}
}

\end{document}